\documentclass[letterpaper]{article} 
\usepackage{aaai23_arxiv}  
\usepackage{times}  
\usepackage{helvet}  
\usepackage{courier}  
\usepackage{graphicx} 
\usepackage{natbib}  
\usepackage{caption} 
\usepackage{algorithm}
\usepackage{algorithmic}
\usepackage[absolute,overlay]{textpos}

\usepackage{newfloat}
\usepackage{listings}

\usepackage{epigraph}
\usepackage{amssymb}
\usepackage{booktabs}
\usepackage{siunitx}
\usepackage{colortbl}
\usepackage{multirow}
\usepackage{amsmath}
\usepackage{placeins}
\usepackage{pifont}
\usepackage[dvipsnames,table]{xcolor}
\usepackage{colortbl}
\usepackage[colorlinks]{hyperref}
\usepackage{soul}
\definecolor{MK_Three_One}{RGB}{118,42,131}
\definecolor{MK_Three_Two}{RGB}{175,141,195}
\definecolor{MK_Three_Three}{RGB}{231,212,232}
\definecolor{MK_Three_Four}{RGB}{217,240,211}
\definecolor{MK_Three_Five}{RGB}{127,191,123}
\definecolor{MK_Three_Six}{RGB}{27,120,55}
\hypersetup{
 linkcolor=MK_Three_One
,citecolor=MK_Three_Two
,filecolor=MK_Three_Three
,urlcolor= MK_Three_Six
,menucolor=MK_Three_Five
,runcolor=MK_Three_Four
,linkbordercolor=MK_Three_One
,citebordercolor=MK_Three_Two
,filebordercolor=MK_Three_Three
,urlbordercolor=MK_Three_Six
,menubordercolor=MK_Three_Five
,runbordercolor=MK_Three_Four
}

\DeclareCaptionStyle{ruled}{labelfont=normalfont,labelsep=colon,strut=off} 
\floatstyle{ruled}
\newfloat{listing}{tb}{lst}{}
\floatname{listing}{Listing}
\title{ExpeL: LLM Agents Are Experiential Learners}
\author {
    Andrew Zhao,\textsuperscript{\rm $\spadesuit$}
    Daniel Huang, \textsuperscript{\rm $\clubsuit$}
    Quentin Xu, \textsuperscript{\rm $\clubsuit$}
    Matthieu Lin, \textsuperscript{\rm $\clubsuit$}
    Yong-Jin Liu, \textsuperscript{\rm $\clubsuit$}
    Gao Huang \textsuperscript{\rm $\spadesuit$}\thanks{Corresponding author.}
}
\affiliations {
    \textsuperscript{\rm $\spadesuit$} Department of Automation, BNRist, Tsinghua University\\
    \textsuperscript{\rm $\clubsuit$} Department of Computer Science, BNRist, Tsinghua University\\
    \texttt{\{zqc21,huang-jy22,xgd22,lyh21\}@mails.tsinghua.edu.cn},\\
    \texttt{\{liuyongjin,gaohuang\}@tsinghua.edu.cn}
}

\begin{document}

\maketitle

\begin{figure}[!ht]
    \centering
    \includegraphics[width=0.254\textwidth]{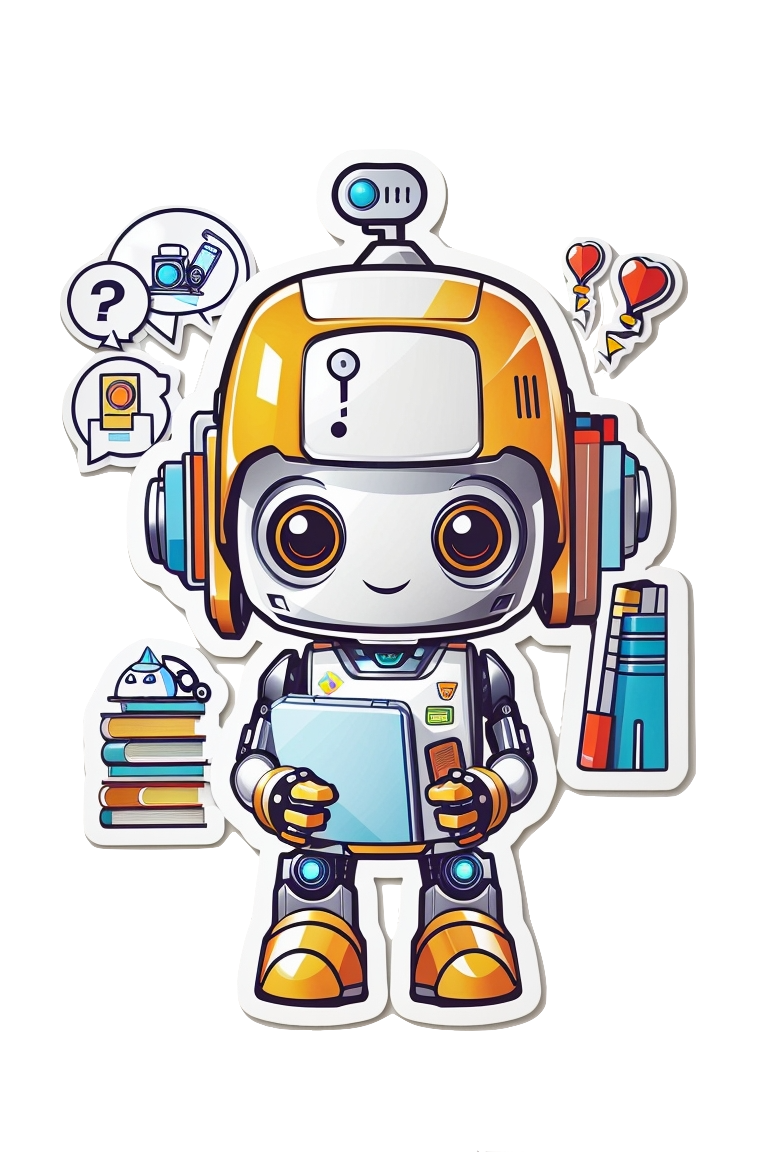}\hfill
    \includegraphics[width=0.254\textwidth]{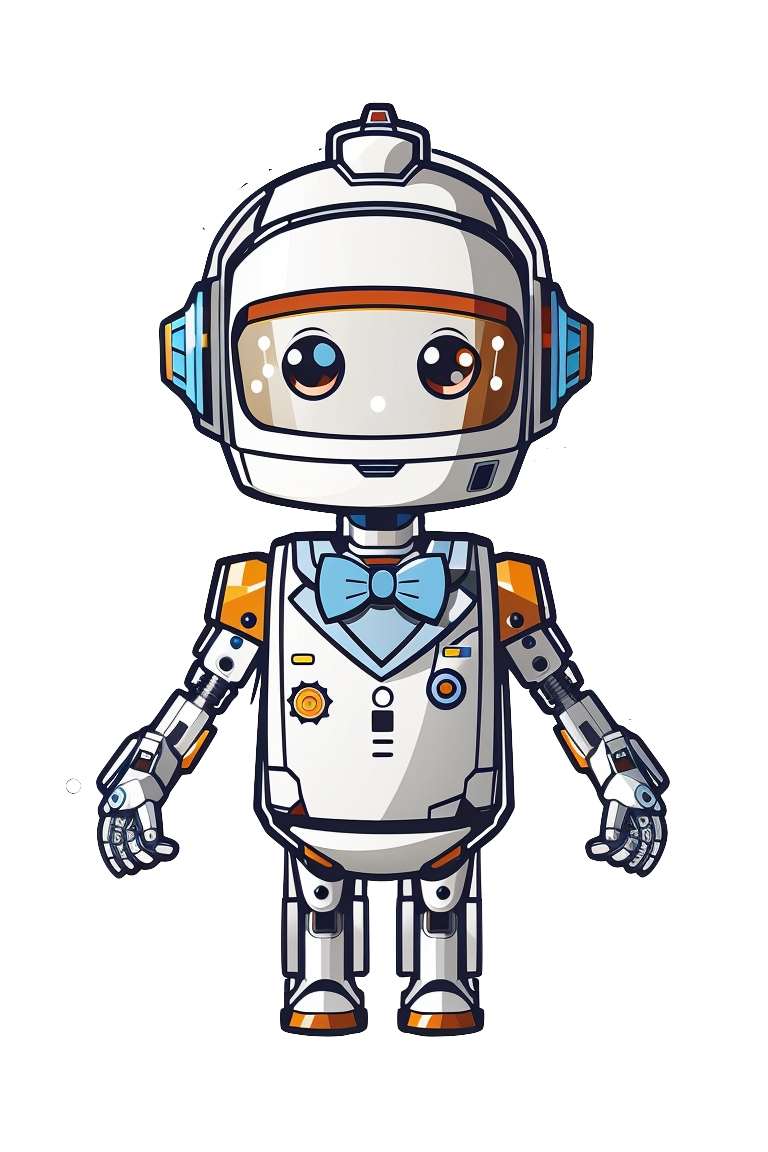}\hfill
    \includegraphics[width=0.254\textwidth]{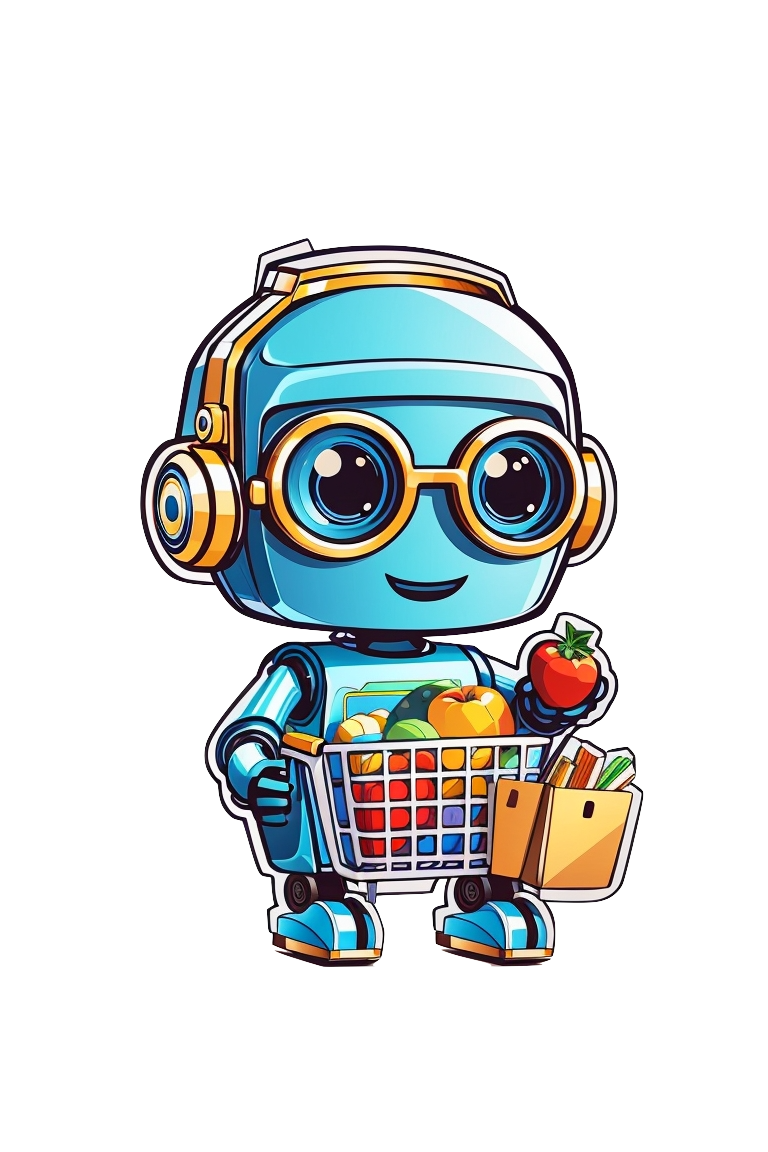}
\end{figure}
\begin{abstract}
The recent surge in research interest in applying large language models (LLMs) to decision-making tasks has flourished by leveraging the extensive world knowledge embedded in LLMs. While there is a growing demand to tailor LLMs for custom decision-making tasks, finetuning them for specific tasks is resource-intensive and may diminish the model's generalization capabilities. Moreover, state-of-the-art language models like GPT-4 and Claude are primarily accessible through API calls, with their parametric weights remaining proprietary and unavailable to the public. This scenario emphasizes the growing need for new methodologies that allow learning from agent experiences without requiring parametric updates. To address these problems, we introduce the Experiential Learning (ExpeL) agent. Our agent autonomously gathers experiences and extracts knowledge using natural language from a collection of training tasks. At inference, the agent recalls its extracted insights and past experiences to make informed decisions. Our empirical results highlight the robust learning efficacy of the ExpeL agent, indicating a consistent enhancement in its performance as it accumulates experiences. We further explore the emerging capabilities and transfer learning potential of the ExpeL agent through qualitative observations and additional experiments.\footnote{Visit \url{https://andrewzh112.github.io/expel} for project page, and \url{https://github.com/LeapLabTHU/ExpeL} for code.}
\end{abstract}

\epigraph{A computer program is said to \textbf{learn} from \textbf{experience} $E$ with respect to some class of tasks $T$ and performance measure $P$, if its performance at tasks in $T$, as measured by $P$, improves with \textbf{experience} $E$.}{\textit{Tom Mitchell}}

\twocolumn

\begin{figure*}[ht]
    \centering
    \includegraphics[width=\linewidth]{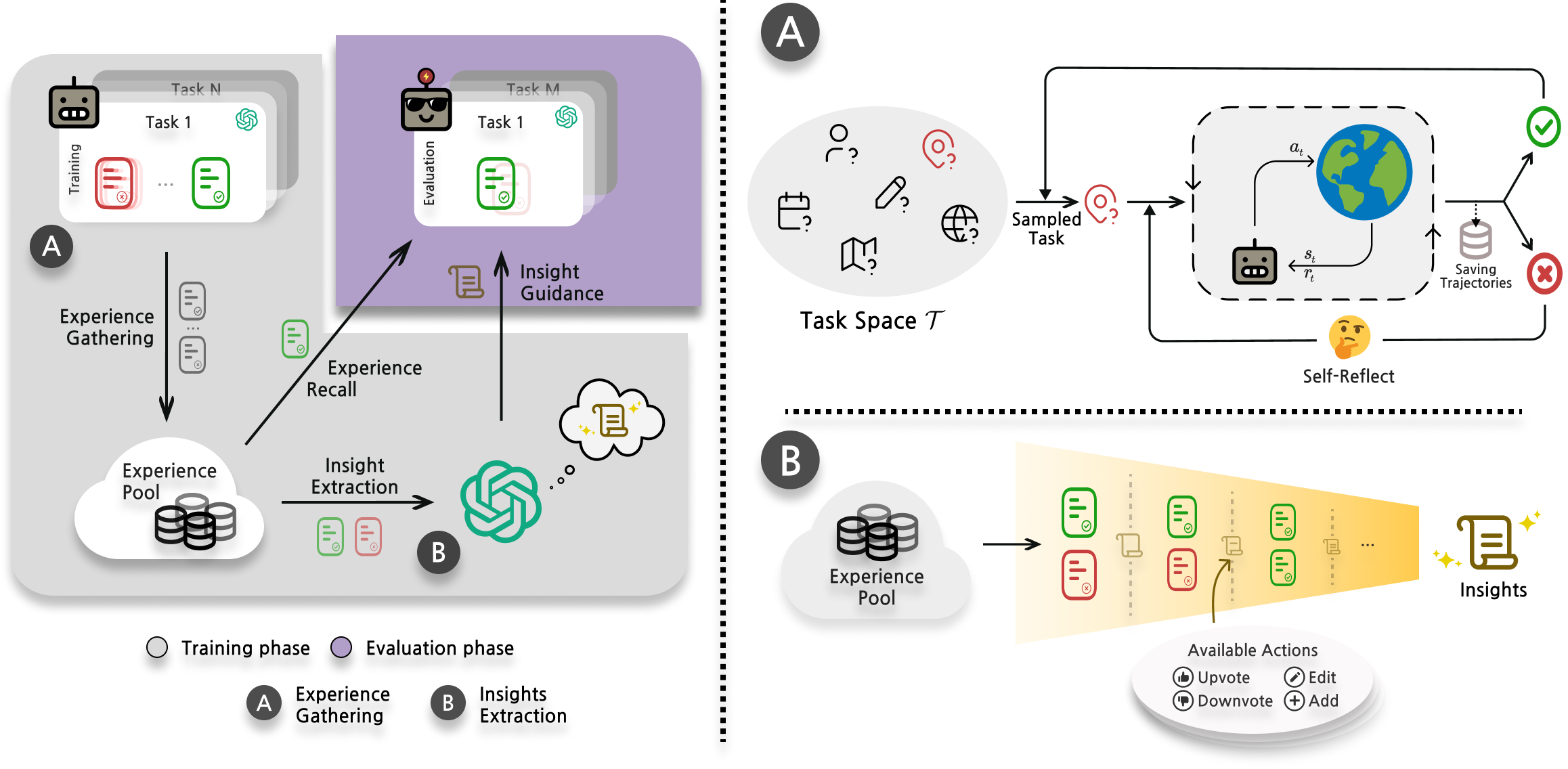}
    \caption{\textbf{ExpeL Agent Overview.} \textit{Left:} ExpeL operates in three stages: (1) Collection of success and failure experiences into a pool. (2) Extraction/abstraction of cross-task knowledge from these experiences. (3) Application of the gained insights and recall of past successes in evaluation tasks. \textit{Right:} (A) Illustrates the experience gathering process via Reflexion \cite{shinn2023reflexion}, enabling task reattempt after self-reflection on failures. (B) Illustrates the insight extraction step. When presented with success/failure pairs or a list of $L$ successes, the agent dynamically modifies an existing list of insights $\hat{\iota}$ using operations \texttt{ADD}, \texttt{UPVOTE}, \texttt{DOWNVOTE}, and \texttt{EDIT}. This process has an emphasis on extracting prevalent failure patterns or best practices.}
    \label{fig:main}
\end{figure*}

\section{Introduction}
Machine learning research has long been captivated by the potential of autonomous agents and their capabilities. In recent times, incorporating large language models into these agents \cite{wang2023survey,xi2023rise} has unveiled a broad spectrum of applications, even extending beyond academia \cite{yang2023foundation,Nakajima2023,SignificantGravitas2023}. One of the significant advantages of LLMs lies in their world knowledge, allowing them to be inherently versatile across various scenarios \cite{zhao2023survey}.

On the one hand, previous works investigated finetuning LLMs with a large number of environment interactions \cite{yao2023retroformer} or with a large amount of human-labeled datasets \cite{nakano2021webgpt,shaw2023pixels}. This class of methods incurs high computational costs and needs access to the LLM's parametric weights. Furthermore, finetuning an LLM restricts its functionalities and can hurt its generalization abilities \cite{du2022shortcut}. On the other hand, prompting methods can augment an LLM with better sequential decision-making planning abilities with only a few in-context examples \cite{hao2023reasoning,lin2023text2motion,sun2023adaplanner}. However, since current LLMs are bounded by context window size \cite{tworkowski2023focused}, these agents have no recollections of what they have seen, and therefore no learning can be done outside of a few demonstrations. \textit{So, how can we strike a balance between these paradigms?}

We present the Experiential Learning (ExpeL) agent as a solution. Our agent autonomously gathers experiences from a collection of training tasks through trial and error. From these experiences, it derives natural language insights and employs its own successful experiences as in-context examples during test time. Our agent's learning process is analogous to a student studying for an exam and then taking it on a single attempt, reflecting many real-world situations. Unlike self-improvement methods like Reflexion \cite{shinn2023reflexion}, our approach emphasizes the importance of retaining experiences across multiple tasks to enhance agent performance. Moreover, ExpeL learns without parameter updates, making it compatible with powerful closed-source models like GPT-4 or Claude. Lastly, the experience-gathering step does not require a large amount of data or human labels.

We evaluated ExpeL on three vastly different domains and consistently outperformed strong baselines. Additionally, we showcased a transfer learning scenario where our agent that accumulated knowledge from source tasks showed positive forward transfer to target tasks. Finally, we highlighted some unexpected emerged abilities the ExpeL agent gained.

In summary, our key contributions are as follows: (1) we introduced ExpeL, a novel LLM agent that autonomously learns from experience  \textit{without gradient updates}; (2) We evaluated ExpeL on a diverse set of tasks to showcase its learning abilities and improvement on top of existing planning methods; (3) we showed a novel setting of transfer learning for our LLM agent and demonstrated forward transferability from source tasks to target tasks. Lastly, we believe that as planning algorithms and foundational models continue to improve, ExpeL's paradigm stands to gain significant benefits from their enhanced performances.

\section{Related Work}
We discuss the most relevant related works in this section. See Appendix \ref{sec:append_related} for detailed discussions on related works.

\textbf{Prompt-based Learning:} Prompt-based learning refines label prediction tasks by modifying the input context, facilitating swift adaptation to new tasks with minimal data \cite{liu2023pre}. This approach capitalizes on LLMs for answers without parameter tuning as they can be augmented using in-context learning \cite{brown2020language}. LAMA \cite{petroni2019language} and GPT-3 \cite{brown2020language} are early works that promoted this formulation. Efforts to reduce the intricacies of prompt design include automatic reasoning chains for NLP \cite{kojima2022large,zhang2022automatic}. Similarly, the ExpeL agent also autonomously learns from experiences using extracted insights and self-generated in-context trajectories by altering the execution prompt.

\noindent\textbf{Retrieval Augmented Generation (RAG):} Retrieval allows LLMs to access databases, mitigating hallucinations \cite{li2022survey,wang2023learning,rubin2021learning,liu2021makes}. Retrieval has also been used to enhance the capabilities of decision-making agents  \cite{humphreys2022large,zhao2023augmenting}. In contrast to these works, we focus on retrieving the ExpeL agent's self-generated experiences, thus reducing the dependency on gold examples and leveraging domain-specific corpus.

\noindent\textbf{Planning for LLM Agents:} Application of LLM agents in fields like robotics, natural sciences, game-playing, and workflows has surged, with emphasis on their world knowledge in fewshot settings \cite{ha2023scaling, mu2023embodiedgpt, bran2023chemcrow, boiko2023emergent, yang2023mm,lin2023swiftsage,nakano2021webgpt,wang2023avalon,liu2023agentbench}. Moreover, LLMs have demonstrated promising zero/few-shot planning and reasoning capabilities in various configurations \cite{sumers2023cognitive}, including embodied environments and reasoning tasks \cite{huang2022language, yao2023tree,wei2022chain, yao2022react,gong2023mindagent}.

\noindent\textbf{Self-improvement and Memory for LLM Agents:} Agents like Reflexion showcase feedback-based improvement, yet often lack cross-task memory \cite{shinn2023reflexion}. Other agents exhibit potential in persistent memory within multi-agent contexts \cite{park2023generative, fable2023showrunner}. Our ExpeL agent combines these approaches, focusing on task-solving while benefiting from self-generated in-context examples and abstracted insights from memory.

\section{Preliminaries}
\subsubsection{Complex Interactive Tasks}
We work with complex interactive tasks where at each time step $i\in\{0,\dots,H\}$, the agent receives an observation $o\in\mathcal{O}$, and from its observation history $H_t$ decides to perform action $a\in\mathcal{A}$. The objective of the agent is to achieve some goal $g\in\mathcal{G}$. We only deal with deterministic environments in this work.

\subsubsection{Large Language Models}
A large language model is a statistical model of the natural language, typically a neural network. In our setting, we use an autoregressive language model \cite{openai2023gpt4,brown2020language,touvron2023llama2,touvron2023llama,chowdhery2022palm}, which given an ordered list of existing tokens $\mathbf{x}=\{x_1,x_2,...,x_{l-1}\}$, outputs the probability of the next token $p(x_l\mid x_{<l})$. An instruction-following LLM \cite{thoppilan2022lamda,chung2022scaling,wei2021finetuned} is typically finetuned on various NLP tasks that are formatted into instruction, input, response tuples \cite{alpaca}. Instruction-tuned models are better at following natural language instructions which alleviates the need for heavy prompt engineering \cite{wei2021finetuned}.

\subsubsection{ReAct and Reflexion}
ReAct \cite{yao2022react} and Reflexion \cite{shinn2023reflexion} are promising frameworks enabling the aforementioned proficiency of LLMs in reasoning and self-improvement. ReAct explicitly intertwines observations, actions, and thoughts, providing a foundation for robust planning and reasoning capabilities. Building upon it, Reflexion introduces an additional reflective step before reattempting the subsequent trial of the same task, enhancing the model's adaptive learning process.

\section{ExpeL: An Experiential Learning Agent}
Recent advancements in generative LLMs suggest an intriguing approach. Rather than altering the LLM parameters, adjusting the prompts may be more beneficial: this strategy ensures that the LLM's inherent common sense knowledge remains intact, allowing for superior generalization \cite{liu2023pre}. Furthermore, some of the most potent language models are proprietary \cite{openai2023gpt4,anthropic2023}. Thus, focusing on prompt-based methods seems promising as a way to harness the strengths of these advanced LLMs. Additionally, previous works on learning in LLM agents have primarily been trained on extensive human-labeled datasets \cite{lin2023swiftsage,shaw2023pixels} or improved via iterative retries \cite{shinn2023reflexion} on a single task. A relatively less explored area is facilitating agents to learn autonomously from their own experiences, similar to a student gaining insights from practicing for an exam. The student tackles practice problems multiple times to derive insights. At the exam, the student rely solely on these insights and draw memories of similar problems to answer the questions with one attempt. With this in mind, we wish to design an LLM agent that autonomously gathers experiences and extracts insights, then uses these cross-task insights and memories of similar tasks to aid its decision-making.

We aim to enhance a planning LLM agent, such as ReAct, with learning abilities that allow it to improve through inter-task experiences without any parameter updates. Inspired by the cognitive abilities inherent in human learning, as well as the benefits observed in self-learning autonomous agents and the progress made in prompt-based methods, we developed the Experiential Learning (ExpeL) agent. During the training stage, the agent interacts with the environment, gathering experiences via trial and error. These experiences are stored in an experience pool \cite{lin1992self}. From this pool, the agent later extracts insights, similar to off-policy learning \cite{watkins1992q}, in which the agent can learn from experiences of a behavior policy. During the evaluation stage, the agent attempts unseen tasks with a single try, augmented with extracted insights and successful trajectories in its experience pool gathered from the training stage. Refer to Fig. \ref{fig:main} for detailed information on our agent framework.

\subsection{Gathering Experiences}
To gather diverse experiences that can be useful to extract information from, we leverage Reflexion \cite{shinn2023reflexion} to continuously retry the training task at most $Z$ times. In particular, the agent will be given a training task $t_n$ at the $z$-th trial, fewshot examples $F_\text{manual}$ and past reflections $\nu_{n,z}$ (initially, $\nu_{n,0}$ is the empty string). At first, the agent will attempt the task with fewshot examples concatenated with its current trajectory $\tau_{n,0}$ as the context, and use ReAct \cite{yao2022react} as the base planning algorithm, $\text{LLM}_\text{ReAct}(\cdot\mid\tau_{n,0},F_\text{manual},\nu_{n,0})$. On the $z$-th trial, when the agent finishes the task or the maximum number of steps $H$ is reached, the ExpeL agent's experience pool $\mathcal{B}$ ingests the trajectory $\tau_{n,z}$. Then, if the agent succeeds, it moves on to the next task. However, if the agent fails, it will look at its failed trajectory and self-reflect to produce $\nu_{n,z+1}=\operatorname{concat}(\nu_{n,z},\text{LLM}_\text{reflect}(\tau_{n,z}))$ to see where it can do better on the next retry, concatenated with the previous reflections. In the next retry, the agent will augment its context with reflection $\nu_{n,z+1}$, the input to the LLM policy, $\text{LLM}_\text{ReAct}(\cdot\mid \tau_{n,z+1},F_\text{manual},\nu_{n,z+1})$.

To highlight, this trial and error way of gathering experiences not only improves the chances of getting more positive examples for experience recall during evaluation but also allows for collecting valuable success/failure pairs used for comparisons during insight extraction (Sec. \ref{sec:insight}). The pseudo-code can be found in Alg. \ref{alg:exp_gather}.

\subsection{Learning from Experiences}
Human learning occurs mainly either by storing successful trajectories in memory, which can be later recalled as specific examples, or by extracting high-level insights from experiences, enabling generalization to novel situations. ExpeL considers both of these learning modes to boost task performance. Concretely, an instruction $I$ given to an LLM agent can be broken down into task specifications and fewshot examples. We can augment task specifications with an agent's extracted insights from past experiences, where an instruction-following LLM can be leveraged \cite{openai2023gpt4} to follow them closely. For fewshot examples, we can allow the agent to retrieve from its experience pool with top-$k$ relevant examples to aid its decisions. Next, we detail our experience recall and insight extraction mechanisms.

\subsubsection{Similar Experiences as Demonstrations}
\label{sec:recollection}

Works have shown that using in-context examples that are semantically similar to the task at hand results in better performance \cite{liu2021makes}. Moreover, when involved in a novel situation, humans also recall from their memory similar tasks they've solved as references when attempting the task \cite{kahneman2011thinking}. Motivated by these observations, we propose experience recall to retrieve successful trajectories from the experience pool gathered during training based on task similarity.

Concretely, we used the \texttt{Faiss} vectorstore \cite{johnson2019billion} as the experience pool, kNN retriever and \texttt{all-mpnet-base-v2} \cite{song2020mpnet} embedder to obtain top-$k$ successful trajectories that have the maximum inner-product task similarity with the evaluation task. The advantage of using task similarity as the retrieval rank is that if the agent repeats a task or does a task similar to an existing successful trajectory from the experience pool, the agent only needs to closely imitate the successful trajectory and have less burden on ability extrapolation.

\subsubsection{Learning from Successes and Failures}
\label{sec:insight}

To leverage the diverse outcomes gathered during the experience collection phase, we believe the agent should analyze experiences in two distinct ways. First, we let the agent compare a failed trajectory with a successful trajectory for the \textit{same} task. This comparison offers a concrete understanding of the agent's shortcomings, highlighting the correct and incorrect actions. Second, we let the agent identify patterns within a set of successful trajectories from \textit{different} tasks. This approach sheds light on common ``good practices" that the agent can adopt to ensure success in evaluation tasks.

\begin{figure}[htbp] 
    \centering
    \includegraphics[width=0.998\linewidth]{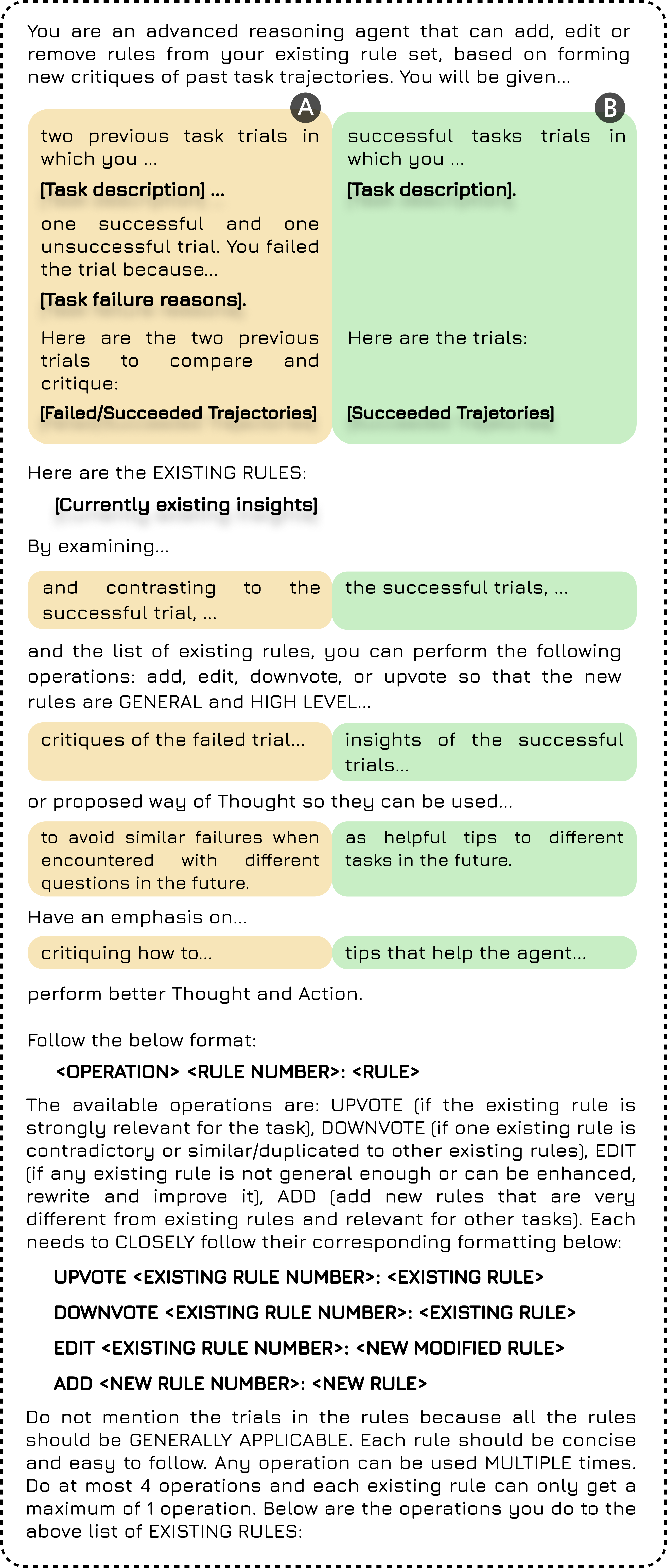}
    \caption{\textbf{Insight Extraction Prompt Template.} The prompt template ExpeL agents used for insight extraction. The same template is used both for success/fail pairs (A, in yellow) and $L$-sized successes (B, in green).}
    \label{fig:insight_prompt}
\end{figure}

For the implementation, we give the agent's instruction-following $\text{LLM}_\text{insights}$ several operators to apply on an existing set of insights $\hat{\iota}$. We initialize the set of insights to an empty set $\hat{\iota}=\emptyset$ and iteratively provide the LLM with fail/success pairs or lists of $L$ successes (created by sampling without replacement) from the experience pool. The operations the LLM can perform are: \texttt{ADD} a new insight, \texttt{EDIT} the content of an existing insight, \texttt{DOWNVOTE} to disagree with an existing insight, or \texttt{UPVOTE} to agree with an existing insight. A newly added insight will have an initial importance count of two associated with it, and the count will increment if subsequent operators \texttt{UPVOTE} or \texttt{EDIT} are applied to it and will decrement when \texttt{DOWNVOTE} is applied to it. If an insight's importance count reaches zero, it will be removed. This particular design choice robustifies the process since even successful trajectories can be suboptimal and mislead the generated insights. The prompt template we used can be found in Fig. \ref{fig:insight_prompt}. We kept the maximum size for a list of successes to $L$ and used \texttt{gpt-4-0613} as the default $\text{LLM}_\text{insights}$. We \textit{empirically} found that \texttt{gpt-4-0613} is better than \texttt{gpt-3.5-turbo-0613} at following instructions on how to use the insight extraction operators and hallucinated less. Pseudo-code for this process can be found in Alg. \ref{alg:insight}. Finally, ExpeL utilizes these generated insights $\hat{\iota}$ in the task inference phase, described next.

\subsection{Task Inference}
\begin{figure}[ht]
    \centering
    \includegraphics[width=\columnwidth]{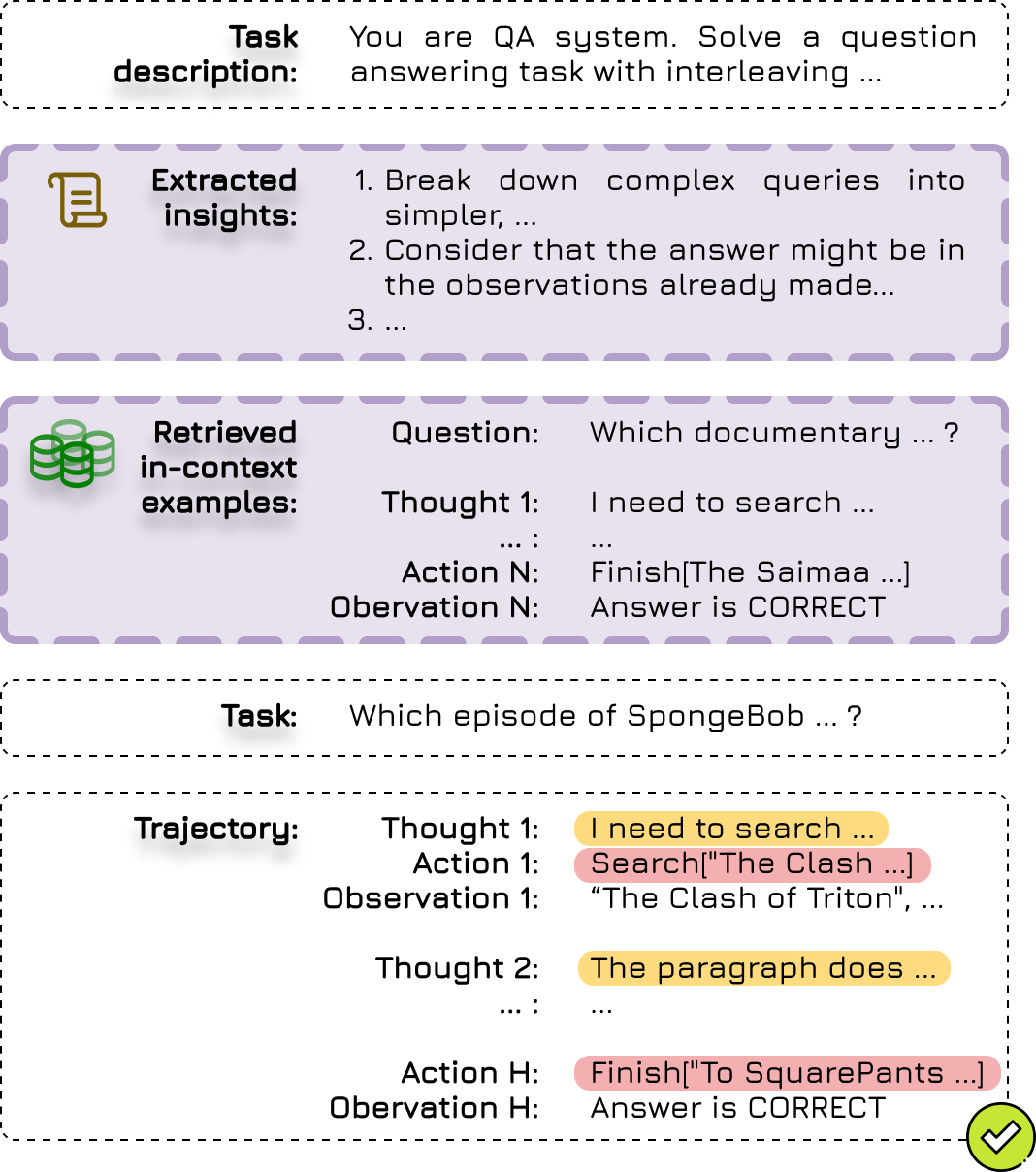}
    \caption{\textbf{Task Inference Prompt Template.} We illustrate ExpeL's prompt template during evaluation. The areas with a white background are identical to the base ReAct agent's inputs. We differ by (purple areas) having additional \textit{extracted insights} from past experience, and dynamically retrieved successful \textit{in-context examples} from past experiences based on task similarity.}
    \label{fig:eval_example}
\end{figure}

After the agent gathers experiences, extracts insights from them, and sets up a vectorstore of successful trajectories, it can proceed to the evaluation. For each task, the task specifications will be augmented with the concatenation of the full list of extracted insights $\hat{\iota}=\operatorname{concat}(\iota_1,\iota_2,\iota_3, ...)$, and the top-$k$ trajectories with the highest task similarity will be retrieved and used as fewshot in-context examples, $F_\text{similar tasks}$. Fig. \ref{fig:eval_example} shows an example prompt template structure, and a pseudo-code for this step can be found in Alg. \ref{alg:eval}. We believe as the list of extracted insights grows, retrieval could be a feasible solution to manage the context window size.

\subsection{Transfer Learning}
After demonstrating how learning by using experiences from a training set can benefit an LLM agent in solving an unseen task in the same task distribution, we investigate another interesting setting where knowledge accumulated from a source task distribution could be useful for a target task distribution with minimal target task examples for the ExpeL agent. Like most transfer learning settings, we assume that the source and target tasks exhibit common knowledge. Therefore, experiences accumulated from source tasks can benefit the agent in solving a new set of target tasks.

\begin{algorithm}
    \caption{ExpeL - Experience Gathering}
    \begin{algorithmic}
    \STATE \textbf{Initialize:} 
    \STATE Policy LLM\textsubscript{ReAct}
    \STATE Self-reflection model LLM\textsubscript{reflect}
    \STATE Collection of tasks $\mathcal{T}_\text{train}$
    \STATE Fewshot examples $F_\text{manual}$
    \STATE Experience pool $\mathcal{B} \leftarrow \text{$F_{\text{manual}}$}$
    \STATE Number of training tasks $N$
    \STATE Maximum retry number $Z$
    \STATE Maximum step number $H$
    \STATE Current task index $n\leftarrow 1$
    \WHILE{task $n \leq N$}
        \STATE \( t_n \leftarrow \mathcal{T}_\text{train}[n] \)
        \STATE Reflection \( \nu_{n,0} \leftarrow \text{``"} \)
        \FOR{trial $z = 0$ to $Z$}
            \STATE \( o_0 \leftarrow \text{env.reset($t_n$)} \)
            \STATE Initialize trajectory \( \tau_{n,z} \leftarrow o_0 \)
            \FOR{timestep \( i=0 \) to \( H \)}
                \STATE \( a_i \leftarrow \text{LLM}_\text{ReAct}(a_i\mid \tau_{n,z}, F_\text{manual},\nu_{n,z}) \)
                \STATE \( o_{i+1}, r_{i+1}, \texttt{done} \leftarrow \text{env.step}(a_i) \)
                \STATE \( \tau_{n,z} \leftarrow \tau_{n,z} \cup \{(o_i,a_i,o_{i+1},r_{i+1})\} \)
                \IF{\( \texttt{done} \)}
                    \STATE break
                \ENDIF
            \ENDFOR
            \STATE \( \mathcal{B} \leftarrow \mathcal{B} \cup \tau_{n,z} \)
            \IF{ \( \texttt{done} \) or \( z=Z \) }
                \STATE \( n \leftarrow n+1 \)
                \STATE break
            \ELSE
                \STATE \( \nu_{n,z+1} \leftarrow \text{concat}(\nu_{n,z} + \text{LLM}_\text{reflect}(\tau_{n,z})) \)
            \ENDIF
        \ENDFOR
    \ENDWHILE
    \RETURN \( \mathcal{B} \)
    \end{algorithmic}
    \label{alg:exp_gather}
    \end{algorithm}
    
    \begin{algorithm}
    \caption{ExpeL - Insight Extraction}
    \begin{algorithmic}
    \STATE \textbf{Initialize:} 
    \STATE Experience pool \( \mathcal{B} \) (from Alg. \ref{alg:exp_gather})
    \STATE Insight extraction model LLM\textsubscript{insights}
    \STATE Set of insights \( \hat{\iota} \leftarrow \emptyset\) 
    \STATE Divide the successes in \( \mathcal{B} \) into $L$-sized chunks:
    \STATE \( C_\text{success} = \{ \{\tau_1^\text{success},\tau_2^\text{success},...\tau_L^\text{success}\}, \)
    \STATE \( \quad \quad \quad \{\tau_{L+1}^\text{success},\tau_{L+2}^\text{success},...\tau_{2L}^\text{success}\}, ... \} \)
    \STATE Construct fail/success tuples of the same tasks in \( \mathcal{B} \):
    \STATE \( C_\text{compare} = \{(\tau_1^\text{success},\tau_{1,0}^\text{fail}), (\tau_1^\text{success},\tau_{1,1}^\text{fail}), ...,\)
    \STATE \(\quad \quad \quad \quad (\tau_2^\text{success},\tau_{2,0}^\text{fail}), ... \}\)
    \FOR{each \( c_\text{compare} \) in \( C_\text{compare} \)}
        \STATE \( \hat{\iota} \leftarrow \text{LLM}_\text{insights}(c_\text{compare},\hat{\iota}) \)
    \ENDFOR
    \FOR{each \( c_\text{success} \) in \( C_\text{success} \)}
        \STATE \( \hat{\iota} \leftarrow \text{LLM}_\text{insights}(c_\text{success},\hat{\iota}) \)
    \ENDFOR
    \RETURN \( \hat{\iota} \)
    \end{algorithmic}
    \label{alg:insight}
    \end{algorithm}
    
    \begin{algorithm}
    \caption{ExpeL - Evaluation}
    \begin{algorithmic}
    \STATE \textbf{Initialize:} 
    \STATE ExpeL agent LLM\textsubscript{ExpeL}
    \STATE Text Embedder \( \mathcal{E} \)
    \STATE Experience pool \( \mathcal{B} \) (from Alg. \ref{alg:exp_gather})
    \STATE Set of insights \( \hat{\iota} \) (from Alg. \ref{alg:insight})
    \STATE Collection of evaluation tasks \( \mathcal{T}_\text{evaluation} \)
    \STATE Number of evaluation tasks $M$
    \STATE Number of fewshots $k$
    \STATE Number of successes \( S \leftarrow 0 \)
    \FOR{task \( m = 1 \) to \( M \)}
        \STATE \( t_m \leftarrow \mathcal{T}_\text{evaluation}[m] \)
        \STATE \( o_0 \leftarrow \text{env.reset($t_m$)} \)
        \STATE Initialize trajectory \( \tau_{m} \leftarrow o_0 \)
        \STATE \( F_\text{similar tasks} \leftarrow \texttt{Faiss}(t_m, \mathcal{B}, \mathcal{E}, k) \)
        \FOR{timestep \( i = 1\) to \( H \)}
            \STATE \( a_i \leftarrow \text{LLM}_\text{ExpeL}(a_i\mid \tau_{m}, F_\text{similar tasks}, \hat{\iota}) \)
            \STATE \( o_{i+1}, r_{i+1}, \texttt{done} \leftarrow \text{env.step}(a_i) \)
            \STATE \( \tau_{m} \leftarrow \tau_{m} \cup \{(o_i,a_i,o_{i+1},r_{i+1})\} \)
            \IF{\( \texttt{done} \)}
                \STATE break
            \ENDIF
        \ENDFOR
        \IF{\( r_{i+1} = 1 \)}
            \STATE \( S \leftarrow S + 1 \)
        \ENDIF
    \ENDFOR
    \RETURN \( \frac{S}{M} \)
    \end{algorithmic}
    \label{alg:eval}
    \end{algorithm}

\begin{figure}[ht]
        \centering
        \includegraphics[width=\linewidth]{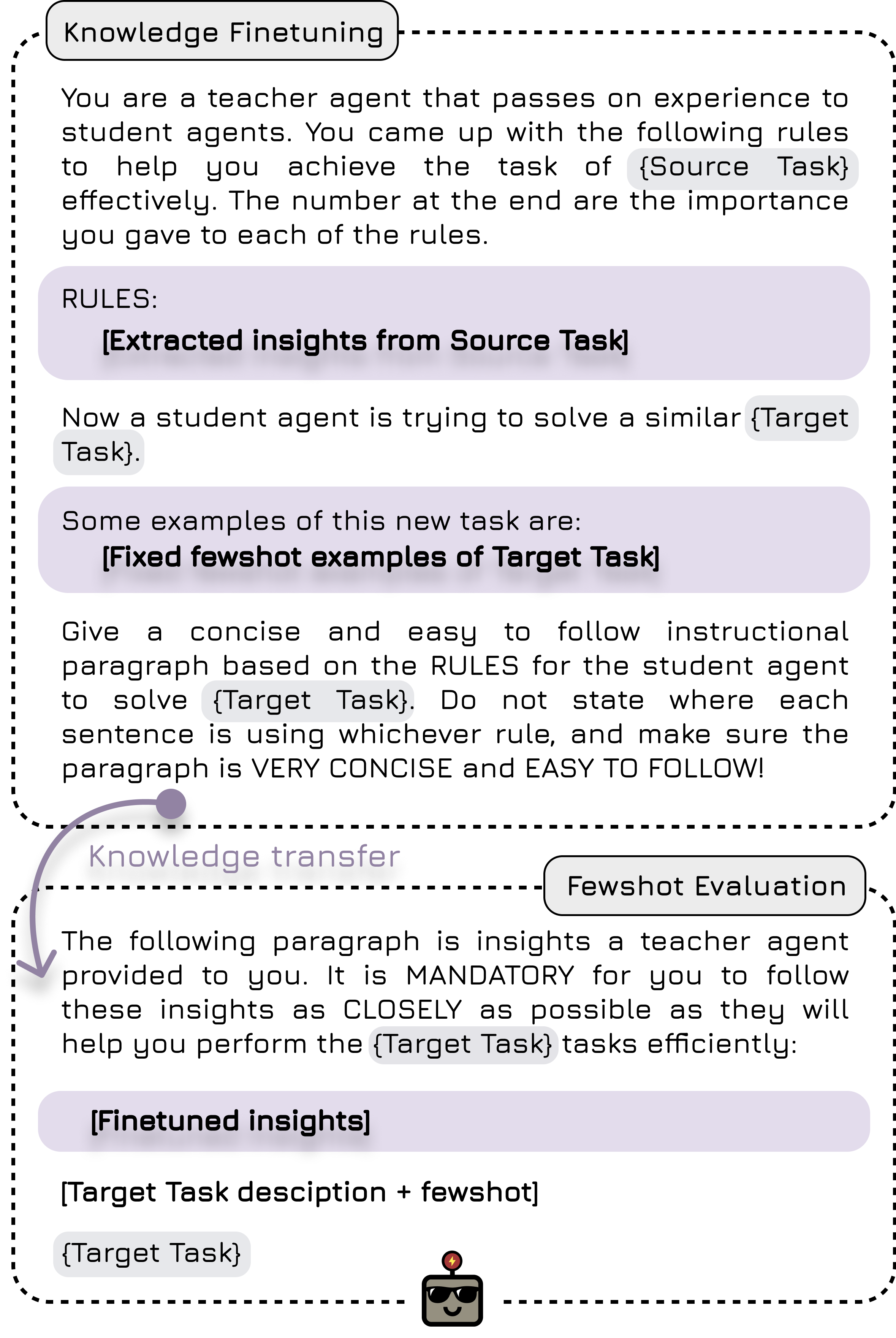}
        \caption{\textbf{Transfer Learning Finetuning Prompt Template.} The prompt template used to finetune knowledge from source to target domain. Highlighted in grey should be formatted with concise descriptions of the tasks.}
        \label{fig:transfer_prompt}
    \end{figure}

    Similar to pretraining on source task and finetuning on target task in transfer learning literature \cite{zhuang2020comprehensive}, we propose to use the extracted insights \( \hat{\iota} \) from the source task and fewshot examples from the target task to ``finetune" the insights so that they are more applicable in the target task. We hypothesize that using target task fewshot examples can better ground the insights into the target task and mitigate hallucinations. An example prompt template to ``finetune" extracted insights from a source domain to tailor them to a target domain is illustrated in Fig. \ref{fig:transfer_prompt}.

\begin{figure*}[ht]
    \centering
    \includegraphics[width=\linewidth]{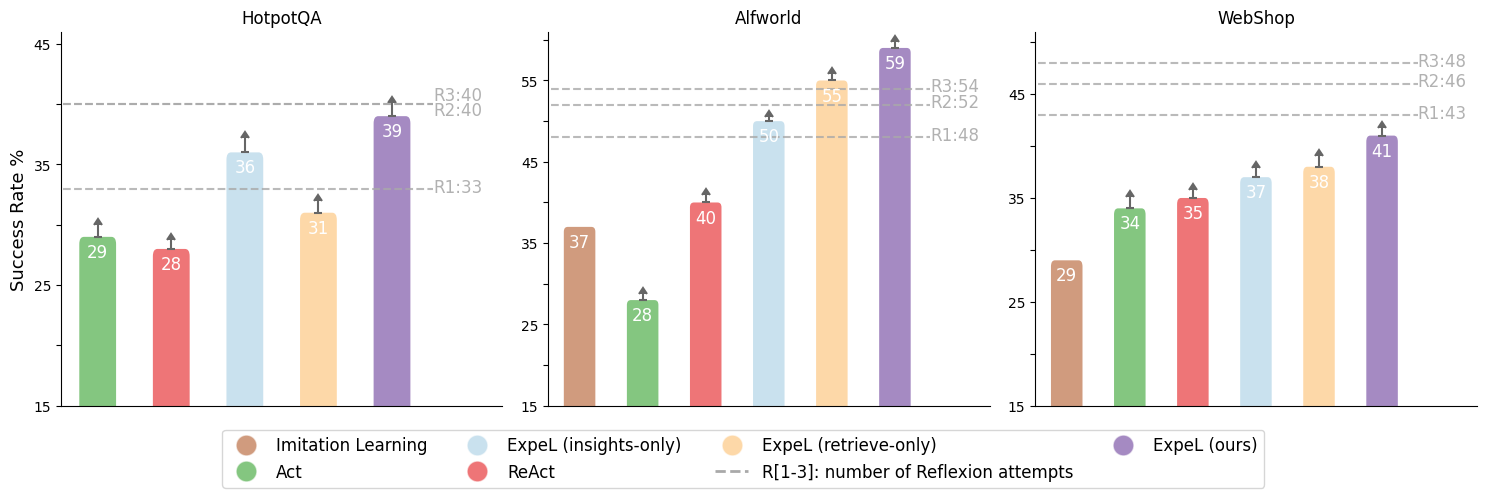}
    \caption{\textbf{Main Results.} Average task success rates (std. error in gray arrows) across three different domains: HotpotQA, ALFWorld, and WebShop. ReAct and Act are used as baselines. ExpeL consistently outperforms the baselines on all domains, highlighting the importance of learning from experience. Additionally, we compare ExpeL with ExpeL (retrieve-only) and ExpeL (insights-only) to highlight that both insight extraction and task similarity retrieval are \textit{essential} and \textit{synergistic}.}
    \label{fig:main_results}
\end{figure*}

\subsection{ExpeL's Strengths}
In this section, we outline the key strengths of our framework. First and foremost, ExpeL offers inherent interpretability, as both the extracted experiences and successful trajectories are presented in natural language. This design allows users to easily inspect, modify, or remove potentially harmful trajectories/insights — a challenge in finetuned models. Moreover, users can seamlessly add expert insights or trajectories to an ExpeL agent. Additionally, our learning approach is highly accessible; it demands less data, reduces computational resources, and is straightforward to implement. Furthermore, self-improvement methods like Reflexion \cite{shinn2023reflexion} facilitate intra-task improvements, but ExpeL enables inter-task learning. ExpeL does not rely on retries during deployment, which certain domains require. On the flexibility front, the ExpeL agent boasts a significant level of versatility. It is not restricted to specific language models and complements existing strategies aimed at enhancing LLM agent planning capabilities. Moreover, when applied in conjunction with them, ExpeL might even improve the capabilities of finetuned agents. Another strength lies in continuous improvement. Our method stands to benefit from the ongoing enhancements in foundational models. As an illustration, our experiments show that using \texttt{gpt-4} to extract insights outperforms \texttt{gpt-3.5-turbo} (refer to Sec. \ref{sec:ablation}). Lastly, we introduced a method for transferring extracted insights across domains using only a small amount of finetuning examples, demonstrating the advantage of our approach in diverse settings with limited data.

\section{Experiments}
\subsection{Experimental Setup}
In line with ReAct \cite{yao2022react}, the experiments are designed based on four text-based benchmarks: HotpotQA \cite{yang2018hotpotqa}, a knowledge-intensive dataset that challenges an agent to perform reasoning and question answering using the search tool Wikipedia Docstore API, ALFWorld and WebShop \cite{shridhar2020alfworld, yao2022webshop} that require the agent to perform interactive multi-step decision-making tasks in respectively a household and an online shopping website environments, and FEVER \cite{thorne2018fever}, that focuses on fact verification tasks using the same API as HotpotQA which makes it suitable for knowledge transfer (Sec. \ref{sec:transfer}). All experiments use four-fold validation, and we report the mean and standard error over the folds. Following ReAct, for all environments, we use success rate as the evaluation metric: exact matching for HotpotQA and FEVER, completing the task in time for ALFWorld, and purchasing the item that matches all attributes for WebShop. Some additional metrics are introduced when the environment offers them: mean reward (calculated using Eq. 1 in Appendix) score $r\in [0,1]$ for WebShop and a score breakdown per task type for ALFWorld.

We use ReAct and Act as main baselines planning LLM agents \cite{yao2022react}, where Act does not have the reasoning steps like ReAct. All agents, including ExpeL, used \texttt{gpt-3.5-turbo-0613} when performing actions during evaluation. All text generations were done with temperature 0 and greedy decoding. Imitation learning (IL) results were taken from the ReAct paper \cite{yao2022react}. More details about the experimental setup can be found in Appendix \ref{sec:env_details}.

\subsection{Main Results}

The primary findings of this study are presented in Fig. \ref{fig:main_results}. IL-based method struggles to efficiently perform in WebShop and ALFWorld, possibly due to their demand for more substantial prior and reasoning abilities, which conventional trainings from scratch fail to provide. This limitation shows the promise of leveraging knowledge-based language models to address these challenges. The following claims were made based on (1) a deep understanding of each environment; (2) extracted insights and retrievable in-context examples; and (3) statistics (e.g. number of invalid actions per trial) of the runs.

\subsubsection{Experiential learning}
Augmenting agents with abstracted insights and the ability to recall successful trajectories improve performance across all environments compared to baseline agents. When restricting the ExpeL agent to only one mode of learning (insights-only or retrieval-only), HotpotQA and ALFWorld environments demonstrate contrasting quantitative distinctions (36\%/31\% and 50\%/55\% for HotpotQA and ALFWorld, respectively). The prominent influence of insights on HotpotQA can be due to its reliance on analysing (Wikipedia results) abilities. This highlights the need for general guidelines across various question types. Conversely, ALFWorld's task completion, dependent on specific action sets, is better derived from past experiential trajectories. Furthermore, WebShop presents a unique challenge, requiring both website-based reasoning (price comparisons, query reformulation, etc.) and precise execution of actions (searching, clicking, option selection, etc.). Consequently, the performance across these tasks shows a near equilibrium, as reflected in both the success rate and score (37\%/38\% and 0.675/0.67 for insights/retrieve-only respectively, see Tab. \ref{tab:main-result} in Appendix for scores). These observations highlight the \textit{synergistic} interplay between \textit{abstraction} and \textit{recollection} in experiential learning, with ExpeL showing a quantitative advantage over baseline/restricted learning mode agents.

\subsubsection{Cross-task learning}
Another important finding we observe is the comparison with the Reflexion agent \cite{shinn2023reflexion}. ExpeL matches Reflexion's performance (40\% at R3 vs. 39\%) for HotpotQA and even outperforms it for ALFWorld (54\% at R3 vs. 59\%) \textit{without repeated attempts}. While Reflexion improves results by iteratively refining insights through repeated task execution (R1, R2, R3...), our ExpeL agent leverages cross-task learning by accumulating task experience. However, it is noteworthy that there remains room for improvement in the context of WebShop tasks, approaching the lower side of Reflexion's success rates.

\subsection{Agent Behavioral Analysis}
\label{sec:emergent}

In this section, we highlight some observations made by manually inspecting the trajectories of ReAct agents and ExpeL agents, and by pinpointing possible causes of how some unexpected behaviors might have emerged. Please visit the paper's webpage, \url{https://andrewzh112.github.io/expel}, for full trajectory demos illustrating the following findings.

\subsubsection{Hypothesis Formulation \& Constraints Adaptation}
After extracting the insights from experiences gathered in the training set, we noticed the agent subsequently gained the ability to \textit{reassess} its whole trajectory in the \textit{last steps} and conclusively end the task rather than expressing its ineptitude in providing a solution. This ability was particularly observed in HotpotQA (Fig. \ref{fig:hotpot_emergent1}, \ref{fig:hotpot_emergent2} in Appendix) where a likely influential insight was stating that the agent should ``consider the answer might be in the observations already made". Therefore the agent would finish by proposing the most probable answer given its past observations rather than concluding with ``Unknown" or ``Information not available".

\subsubsection{World Model Belief Update}
We noticed our ExpeL agent updated its beliefs through the insights and over its gained experience. This belief thereby update enables the agent to avoid unnecessary actions and increase efficiency in solving a given task. For example, in ALFWorld, the agent completely changed the priors it had in ReAct on the likely locations of a pan (from drawers/countertops/cabinets to stoveburners). This behavior emerged from the extracted insight claiming that ``when searching for an item" it needs to ``consider its nature and its typical usage" (Fig. \ref{fig:alf_emergent} in Appendix), leading the agent to promptly and accurately find the correct item at the first step while the ReAct agent could not find it in time.

\subsubsection{Self-correction}
Although ReAct was sometimes not able to reassess its situation when attempting to solve a task, ExpeL demonstrated its proficiency in identifying and rectifying missteps. Notably, when incorrectly taking an object in ALFWorld, the agent has shown its ability to put it back and resume the task by searching for the proper object (Fig. \ref{fig:alf_emergent2} in Appendix). This highlights ExpeL's capacity to recover from errors and stay on course without hallucinating when completing tasks. This behavior is possibly encouraged by the generated insight ``reassess the situation and consider alternative actions" if ``an attempt does not progress the task".

\subsection{Transfer Learning}
\label{sec:transfer}
In this experiment, we use the HotpotQA dataset \cite{yang2018hotpotqa} as source tasks and the FEVER dataset \cite{thorne2018fever} as target tasks. Like the HotpotQA dataset, we equip the agent with the ability to navigate on Wikipedia using a Docstore API; therefore, we hypothesize that some of the knowledge obtained from HotpotQA tasks should also be beneficial when transferred to the FEVER tasks. We use \texttt{gpt-4-0613} for adapting the HotpotQA insights into FEVER insights. We use the same fewshot examples to finetune the insights as the ones that will be used during task execution. We compare our ExpeL Transfer agent's transfer learning ability with (1) ReAct; (2) Act; and (3) an agent that ``finetunes" insights without task demonstrations. Notice that since source and target tasks are inherently different, we do not have an experience pool to retrieve from; thus, the ExpeL Transfer agents use the existing fixed fewshot examples as in-context examples.

Tab. \ref{tab:fever} showcases the transfer learning results. Both agents that transferred knowledge from the source domain saw performance gains. Notably, the agent with a few in-context examples had a more significant improvement than the one without, indicating the effectiveness of the proposed ``finetuning" method in transfer learning scenarios.

\begin{table}
    \centering
    \scalebox{0.95}{
    \begin{tabular}{lc}
        \hline
        \rowcolor{MK_Three_One!10} & \textbf{FEVER (SR \%)} \\
        \hline
        Act & 58 $\pm$ 0.0 \\
        \rowcolor{MK_Three_One!10} ReAct & 63 $\pm$ 0.4 \\
        ExpeL Transfer w/o Task Demos & 65 $\pm$ 1.7 \\
        \rowcolor{MK_Three_One!10} ExpeL Transfer & \textbf{70} $\pm$ 0.7 \\
        \hline
    \end{tabular}
    }
    \caption{\textbf{Transfer Results.} We transfer insights extracted from HotpotQA to FEVER. \textit{Act} and \textit{ReAct} are baseline agents, \textit{ExpeL w/o Task Demos} does not utilize fewshot examples when altering the insights for the target task.}
    \label{tab:fever}
\end{table}

\begin{table}
    \centering
    \scalebox{0.95}{
    \begin{tabular}{lcccc}
        \hline
        \rowcolor{MK_Three_One!10} 
            & \textbf{R0} & \textbf{R1} & \textbf{R2} & \textbf{R3} \\ \hline
        ReAct+Reflexion &  40.3\%        & 47.8\%               & 52.2\%               & 54.4\%             \\ 
        \rowcolor{MK_Three_One!10} 
        ExpeL retrieve only &  54.5\%        & 57.5\%               & 59.7\%               & 60.4\%             \\
        ExpeL+Reflexion          & \textbf{\hl{59.0\%}}         & \textbf{60.4\%}             & \textbf{63.4\%}                 &\textbf{64.2\%}                \\
        \hline
    \end{tabular}
    }
    \caption{\textbf{Success Rate on ALFWorld with Reflexion Rounds.} ExpeL and Reflexion appear to be synergistic in the ALFWorld environment (\hl{Highlight} = ExpeL with one attempt). R1-R3 were obtained from failed R0 checkpoints.}
    \label{tab:reflexion}
\end{table}

\subsection{ExpeL with Task Reattempts}
While not being the central focus of our study, we present preliminary findings on the effectiveness of incorporating task reattempts into the evaluation phase using ExpeL by resuming the failed checkpoints from R0. The performance of ExpeL combined with Reflexion, alongside two baselines: ReAct/Reflexion and ExpeL without insights (ExpeL retrieve only), is detailed in Table \ref{tab:reflexion}. The results demonstrate a notable improvement in the success rate when ExpeL is paired with Reflexion, with the success rate increasing as the number of task reattempts grows.

\subsection{Ablation Studies}
\label{sec:ablation}

\begin{figure}[ht]
    \centering
    \includegraphics[width=\columnwidth]{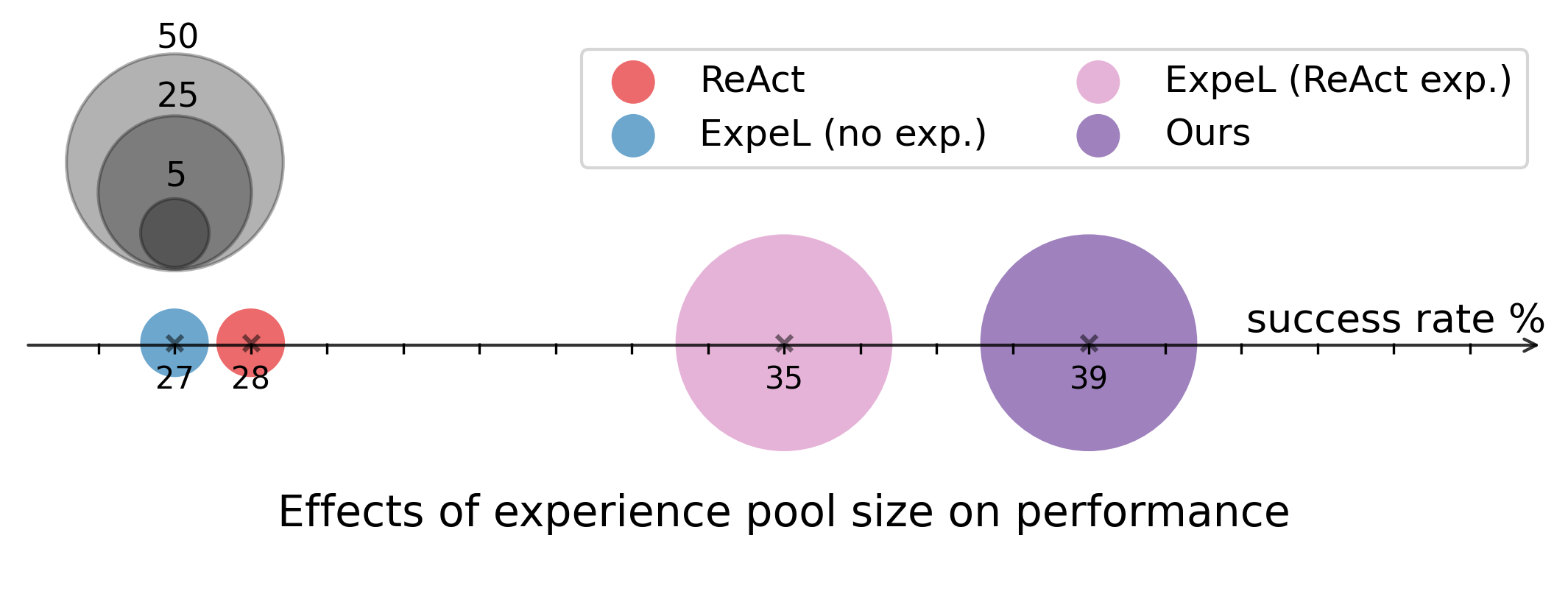}
    \caption{\textbf{Effects of Experience on Performance.} We highlight the correlation between the number of \textit{diverse} experience samples and the final performance. Concretely, we compare ExpeL with (1) ReAct, (2) ExpeL that only has access to fewshot examples, and (3) ExpeL that only uses ReAct during the experience gathering step. It is evident that extra autonomously collected experiences are essential to ExpeL's success and that diversity of success/failure data gathered using Reflexion was superior to using ReAct only.}
    \label{fig:experience_ablation}
\end{figure}

One main component of ExpeL is the agent's ability to autonomously gather valuable experiences benefiting its own learning. Therefore, we wish to investigate if the number of \textit{useful} experiences impacts the downstream performance of ExpeL. We designed two different agents to compare our agent with. The first one only has access to initial fewshot examples and extracts insights from them. The second gathers experience using ReAct where the agent has no retries. Thus, the agent will not only get less successful trajectories but will also not have any success/failure comparison pairs during insights extraction. We conducted experiments in the HotpotQA environment and presented the results in Fig. \ref{fig:experience_ablation}. As we can see, the agent that extracts insights from the existing fewshots has no advantage compared to the ReAct agent, illustrating that experience is essential for ExpeL to learn from. This was reflected in a significantly better performance for the two other agents having access to more experience. Furthermore, the ExpeL agent with access to a \textit{diverse} set of experiences (failure and success pairs obtained using Reflexion) performs better than the agent using only ReAct during experience gathering.

Next, we will scrutinize the efficacy of the insight extraction step of ExpeL. Since insights had the most significant impact on the HotpotQA environment (Fig. \ref{fig:main_results}), we performed the ablations on insights in this environment. We use three dimensions to ablate the design choices for insight extraction by creating the following variants of ExpeL agents: (1) human-crafted insights (Fig. \ref{fig:handcraft} in Appendix), which were manually engineered by carefully studying the agent's mistakes during the experience gathering step; (2) adding reflections $\nu$ into the insights construction step in addition to using fail/success pairs and lists of successes; (3) using \texttt{gpt-3.5-turbo-0613} as the $\text{LLM}_\text{insights}$. Results in Tab. \ref{tab:ablation} show several significant findings: (1) learned insights by the agent are more advantageous than hand-crafted ones; (2) using reflections in addition to success/failure pairs and lists of successes is disadvantageous, possibly due to reflections sometimes outputting hallucinations, therefore misleading the insight extraction stage; and (3) a better LLM is more advantageous at improving ExpeL's performance, suggesting our agent will enjoy free performance boosts with the ever-improving nature of base foundation models.

Lastly, we investigated the design choice of using task similarity as the ranking score for retrieving successful in-context examples in ALFWorld. In particular, we use (1) reason similarity by retrieving top-$k$ trajectories with the most similar reasoning step as the latest reasoning step in the current trajectory, and (2) randomly sampling successful trajectories from the experience pool. We clearly observe in Tab. \ref{tab:ablation} that retrieving with task similarity (ExpeL) performs the best. Reason similarity is still advantageous but slightly drops in performance, possibly due to dynamically changing fewshots during a single trajectory, causing instabilities. Lastly, random sampling has a significant drop in performance, suggesting that our design choice of selecting the most pertinent in-context example is advantageous.

\begin{table}[ht]
    \centering
    \begin{tabular}{lc}
      \toprule
      \rowcolor{MK_Three_One!10} & \textbf{HotpotQA (SR \%)} \\
      \midrule
      ReAct & 28.0 $\pm$ 1.4 \\
       \rowcolor{MK_Three_One!10} Hand-crafted insights & 32.0 $\pm$ 1.1 \\
       Insights with reflections & 29.0 $\pm$ 0.4 \\
       \rowcolor{MK_Three_One!10} \texttt{gpt-3.5-turbo} insights & 32.0 $\pm$ 0.4 \\
       ExpeL (ours) & \textbf{39.0} $\pm$ 1.7 \\
       \bottomrule
       \toprule
       \rowcolor{MK_Three_One!10} & \textbf{ALFWorld (SR \%)} \\
       \midrule
       ReAct & 40.0 $\pm$ 0.3 \\
       \rowcolor{MK_Three_One!10} Reasoning similarity & 48.5 $\pm$ 2.1 \\
       Random sampled & 42.5 $\pm$ 0.8 \\
       \rowcolor{MK_Three_One!10} ExpeL (ours) & \textbf{59.0} $\pm$ 0.3 \\
       \bottomrule
    \end{tabular}
    
    \caption{\textbf{Ablations Results.} \textit{Upper:} \textbf{Ablations on insight extraction.} Hand-crafted insights enjoyed a performance boost over ReAct but were less effective than LLM-generated ones. Furthermore, adding reflections to the insight-generating process hurt performance. Lastly, better LLM base models give better insights. \textit{Lower:} \textbf{Ablations on in-context examples selection strategy.} Randomly selected baseline has a significant drop in performance while ranking using reason similarity also has a noticeable dip.}
    \label{tab:ablation}
  \end{table}

\section{Conclusion and Limitations}
\subsubsection{Limitations}
In this work, we investigated tasks with textual observation, which is limiting in real-world scenarios. Thus, incorporating image observations will make our method more generally applicable. Using Vision-Language Models or captioning models to supplement the LLM to enable image observations could be an interesting new avenue of research. Additionally, we investigated the efficacy of our method by using closed-source API LLMs, which can be off-limits in some applications. Exploring LLM agents using open-source LLMs should be another promising future work \cite{zeng2023agenttuning}. Furthermore, since our extracted insights do not exceed the current LLM's token limit, we can fit them into the agent's context window. However, extra retrieval steps for insights might be needed for truly lifelong learning agents to ensure a manageable context window size. Lastly, unlike reinforcement learning methods, prompting techniques lack theoretical underpinnings that could potentially impact the efficiency of the resulting policies. Future research should explore the integration of these approaches to yield more effective and optimal solutions.

In summary, we introduced ExpeL, a novel learning LLM agent that autonomously gathers experience from a set of training tasks to improve its abilities in solving evaluation tasks without access to model parameters. We demonstrated its learning abilities by showing its performance gain compared to vanilla ReAct and Act agents. Furthermore, we investigated a transfer learning scenario where extracting insights from a set of source tasks can benefit the ExpeL agent in solving a target task. Lastly, we presented several unexpected emerged abilities our agent developed at the end of its training. We believe that autonomously learning from experience is essential for developing human-like intelligent agents, and our ExpeL agent is a step toward that goal.

\section*{Acknowledgement}
This work is supported in part by the National Key R\&D Program of China
(2022ZD0114900), the National Natural Science Foundation of China under Grants 62022048, U2336214, and 62332019, and the Guoqiang Institute of Tsinghua University.

\bibliography{main}

\onecolumn

\appendix
\section*{Appendix}

\section{Detailed Related Works}
\label{sec:append_related}
\subsection{Prompt-based Learning}
\label{sec:prompt_based_learning}
Prompt-based learning is a paradigm where the language model that originally outputs the label $\mathbf{y}$ from context $\mathbf{c}$ improves on the label prediction task with a modified context $\hat{\mathbf{c}}$ \cite{liu2023pre}. This framework is compelling as it enables the usage of pre-trained LLMs trained on vast text volumes. Furthermore, a new prompting function supports fewshot or zero-shot learning, thereby adapting swiftly to tasks with minimal or no labeled data. Specifically, tuning-free prompting directly produces answers using a pre-trained language model's prompt without altering its parameters. This method can be enhanced with answered prompts, a strategy termed in-context learning \cite{brown2020language}. Examples include LAMA \cite{petroni2019language}, GPT-3 \cite{brown2020language} and CoT \cite{wei2022chain}. Its benefits include efficiency, no parameter updates, avoidance of catastrophic forgetting, and zero/fewshot setting applicability. However, it demands intricate prompt engineering and domain knowledge expertise to increase accuracy. Works like AutoPrompt and Zero-shot-CoT \cite{kojima2022large,zhang2022automatic} alleviate the burden on the engineer by automatically generating reasoning chains for NLP reasoning tasks. Likewise, ExpeL agent automatically gathers experiences in sequential decision-making tasks, generates its own insights, and uses these insights alongside successful in-context examples to inform its decisions, taking the burdens away from heavy manual prompt engineering and the requirement of expert domain knowledge.

\subsection{Retrieval Augmented Generation}
\label{sec:retrieval_augmented_generation}
Retrieval augmented generation has gained popularity, which is helpful to reduce hallucination and give LLMs access to internal databases \cite{li2022survey}. Several works in the field of NLP demonstrated the efficacy of retrieving in-context examples \cite{wang2023learning,rubin2021learning} from a database of gold demonstrations. On the contrary, our work explores LLM agents retrieving from their own generated experiences, which lessens the burden of the user's engineering efforts and domain expertise.

\subsection{LLM Agents}
\label{sec:llm_agents}
Research involving using LLMs as the ``brain" of an agent has surged in recent years. LLM agents have been instantiated in many areas such as robotics \cite{ha2023scaling,rt22023arxiv,mu2023embodiedgpt,mirchandani2023large,wu2023tidybot}, natural sciences \cite{bran2023chemcrow,boiko2023emergent} and automated workflows \cite{yang2023mm,gur2023real}. Most of these works leverage LLMs' strong common sense knowledge to achieve downstream tasks in a zero or fewshot manner to keep the LLM's strong world knowledge priors. Our ExpeL agent also leverages the powerful world knowledge of LLMs. Concretely, we use LLMs during gathering experience, extracting insights, and downstream execution steps.

\subsubsection{Planning}
\label{sec:planning}
LLMs have demonstrated the ability to plan in embodied environments in a zero-shot manner \cite{huang2022language}. However, many works show that LLMs' planning ability can be further enhanced by improving their reasoning capabilities \cite{yao2023tree,wei2022chain}. The ReAct agent \cite{yao2022react} demonstrates a combination of reasoning and acting. This approach has not only been proven to be superior to agents that only output actions in various scenarios, but also provides insight into what the agent is thinking while acting. Because of its simplicity and effectiveness, we used ReAct as our base planning algorithm.

\subsubsection{Self-improvement}
\label{sec:self-improvment}
A class of methods that leverages LLMs' ability to self-reflect based on feedback from the environment has shown their superiority compared to algorithms that do not have an awareness of doing the task a second time \cite{shinn2023reflexion,liu2023reflect}. In particular, the Reflexion agent \cite{shinn2023reflexion} provides a verbal hypothesis on why a task failed based on the failed trajectory/environment feedback and improved if given a second chance. However, self-reflecting methods assume the tasks are repeatable, and environment feedback is available at test time. Furthermore, self-reflection methods are stateless and cannot learn cross-task insights. Instead, our approach leverages the strengths of Reflexion and uses it to gather more failed/successful trajectories to extract insights from them and perform better at test time. Works like Voyager \cite{wang2023voyager} explored skill learning in specific environments like Minecraft.

\subsubsection{Memory Mechanisms}
\label{sec:memory_mechanisms}
Agents with persistent long-term memory have demonstrated exciting results in multi-agent settings \cite{park2023generative,fable2023showrunner,qian2023communicative}. These works usually have multiple instantiations of generative agents that interact with each other and simulate human societies or fictional settings. In generative agents \cite{park2023generative}, agents have a memory mechanism where they can retrieve information based on recency, relevance, and importance, much like how humans sometimes refer to and associate with different memories during their day. These lines of work usually are open-ended, while ExpeL agents are task-solving. Like generative agents, our work also uses memory: successful in-context examples and extracted insights as condensed memory which were both gathered from the agent's own experience.

\subsection{Reinforcement Learning}
Our agent gathers experience autonomously, reminiscent of online reinforcement learning methods \cite{sutton2018reinforcement}. Especially, our method uses off-policy learning \cite{watkins1992q}, where the policy uses Reflexion during experience gathering and performs policy improvement via insight extraction and retrieval of similar tasks as in-context examples. Specifically, the retrieval step is similar to experience replay \cite{lin1992self}, where research has been done to select which examples to give the agent for training \cite{schaul2015prioritized,yue2023offline}. However, unlike these existing methods, ExpeL doesn't require access to model parameters, the design of complicated reward or loss functions, or a large number of environment interactions.

\section{Broader Impacts}
\label{sec:broader_impacts}
Our research focuses on LLM agents. If these autonomous programs are given internet access, there's a risk they might cause unexpected harm. However, techniques such as RLHF could potentially mitigate these adverse effects \cite{nakano2021webgpt,ouyang2022training}.

\section{Computational Resources}
\label{sec:computational_resources}
All experiments were performed on a desktop: Intel(R) Core(TM) i9-9900K CPU @ 3.60GHz with 16 cores, 64GB RAM, and a single NVIDIA GeForce RTX 2080 Ti.

\section{Environment Details}
\label{sec:env_details}
\subsection{Evaluation Task Set}
\label{sec:eval_task_set}
We employ four-fold validation for all experiments. We train on one half of the dataset and evaluate on the other half, and vice versa. All results include the mean and standard error of the results across the folds. For HotpotQA, we assess performance using 100 validation tasks from the distractor dev split of the HotPotQA dataset \cite{yang2018hotpotqa}, which were also used by ReAct and Reflexion. In the case of ALFWorld \cite{shridhar2020alfworld}, we utilized the 134 solvable tasks that ReAct and Reflexion used. Similarly, for WebShop tasks, we evaluated using the same 100 tasks used by ReAct and Reflexion.

\subsection{Prompts/Fewshot Examples}
\label{sec:prompts}
We used the same fewshot examples/prompts from ReAct and Reflexion \cite{yao2022react,shinn2023reflexion} during appropriate stages. For WebShop, we added one additional fewshot to make the environment have two fewshot examples. We show our prompt templates in Appendix \ref{sec:example_prompts} and will make the code publicly available.

\subsection{WebShop Environment Specific Detail}
\label{sec:webshop_env_implementation}
We slightly modified WebShop environment found at \url{https://github.com/princeton-nlp/WebShop}. Our goal was to ensure each experiment instantiation was deterministic. In the original version, item prices and price constraints in instructions were generated by sampling from a uniform range. Instead, we used the average value. While this should produce a result similar to the original implementation on average, it ensures consistency across different instantiations for easier reproducibility. Lastly, we extend the items per page from 3 to 10 since recent LLMs saw an increase in context window size that can accommodate more observations.

\subsection{WebShop Reward Function}
\label{sec:webshop_reward_function}
Another metric introduced in the WebShop \cite{yao2022webshop} is their reward function, which converts the similarity between expected product attributes and the attributes of the purchased product into a value ranging from 0 to 1:
\begin{equation}
\begin{split}
r & = \frac{\left|U_{\text{att}} \cap Y_{\text{att}}\right| + \left|U_{\text{opt}} \cap Y_{\text{opt}}\right| + \mathbb{I}\left[y_{\text{price}} \leq u_{\text{price}}\right]}{\left|U_{\text{att}}\right| + \left|U_{\text{opt}}\right| + 1}\cdot r_{\text{type}}, \\
\end{split}
\label{eq:reward_function}
\end{equation}
where,
\begin{equation}
r_{type} = 
\begin{cases}
    0, & \text{if }\text{TextMatch} = 0 \\
    0.1, & \text{if }\text{TextMatch} < 0.1 \\
    0.5, & \text{if }\text{TextMatch} \leq 0.2 \text{ and query not match and category not match} \\
    1, & \text{otherwise.}
\end{cases}
\label{eq:reward_type_function}
\end{equation}
Since a single query could yield multiple appropriate items, WebShop utilizes a matching reward for assessment. The term ``TextMatch" denotes the textual overlap of pronouns, nouns, and proper nouns between the selected product's title and the target product's title \cite{liu2023agentbench}.

\subsection{Base Language Model}
All experiments were conducted using Langchain \cite{chase2023langchain}, making API calls to the OpenAI API. For Reflexion during experience gathering, we used \texttt{gpt-3.5-turbo-0613} and \texttt{gpt-3.5-turbo-16k-0613} when it is over the context window limit. For insight extraction, we used \texttt{gpt-4-0613}. We used \texttt{gpt-3.5-turbo-0613} for all evaluation stage agents. All experiments were conducted from July 10, 2023, to August 10, 2023.

\section{Environment, Agent, Retrieval Parameters}
\begin{table}[ht]
    \centering

    \begin{tabular}{@{}cc@{}}
        \toprule
        \textbf{Retrieval Parameters}   &   \\ 
        \midrule
        Vectorstore & \texttt{Faiss}    \\
        Retriever type  & kNN   \\
        Embedder    & \texttt{all-mpnet-base-v2}  \\
        \midrule
        \textbf{Agent Hyperparameters}  &   \\
        \midrule
        Max Reflection Retries  &   3   \\
        Reflection LLM  & \texttt{gpt-3.5-turbo-0613}   \\
        Policy LLM  & \texttt{gpt-3.5-turbo-0613}   \\
        Insight Extraction LLM  & \texttt{gpt-4-0613}  \\
        Decoding Temperature  & 0  \\
        Decoding Strategy  & greedy  \\
        \midrule
        \textbf{HotpotQA-specific Parameters}  &   \\ 
        \midrule
        Number of Success Examples in Insight Extraction $L$    & 8   \\
        Max Number of Environment Steps $H$     & 7  \\                        
        Max Number of Fewshot Examples $k$    & 6  \\
        Max Number of Reflection Fewshot Examples $k_\text{reflections}$    & 2  \\
        \midrule
        \textbf{WebShop-specific Parameters}   &   \\ 
        \midrule
        Number of Success Examples in Insight Extraction $L$    & 4  \\
        Max Number of Environment Steps $H$ & 15  \\
        Max Number of Fewshot Examples $k$  & 2  \\
        Max Number of Reflection Fewshot Examples $k_\text{reflections}$    & 2  \\
        Searched items per page & 10 \\
        \midrule
        \textbf{ALFWorld-specific Parameters}  &   \\ 
        \midrule
        Number of Success Examples in Insight Extraction $L$    & 8  \\
        Max Number of Environment Steps $H$     & 20  \\
        Max Number of Fewshot Examples $k$    & 2  \\
        Max Number of Reflection Fewshot Examples $k_\text{reflections}$    & 2  \\
        \midrule
        \textbf{FEVER-specific Parameters}  &   \\ 
        \midrule
        Max Number of Environment Steps $H$     & 7  \\
        Max Number of Fewshot Examples $k$    & 3  \\
        \midrule
    \end{tabular}
    \caption{\textbf{Environment, Retrieval and Agent Parameters.}}
    \label{tab:hyperparam}
\end{table} \FloatBarrier

\section{Prompt Templates}
\label{sec:example_prompts}

\subsection{Policy/Actor Prompt Templates}
Policy/actor prompt templates were taken from ReAct \cite{yao2022react} (\url{https://github.com/ysymyth/ReAct}) with minimal alterations to fit extracted insights for our ExpeL agents.
\begin{figure}[ht]
    \centering
    \includegraphics[width=0.94\linewidth]{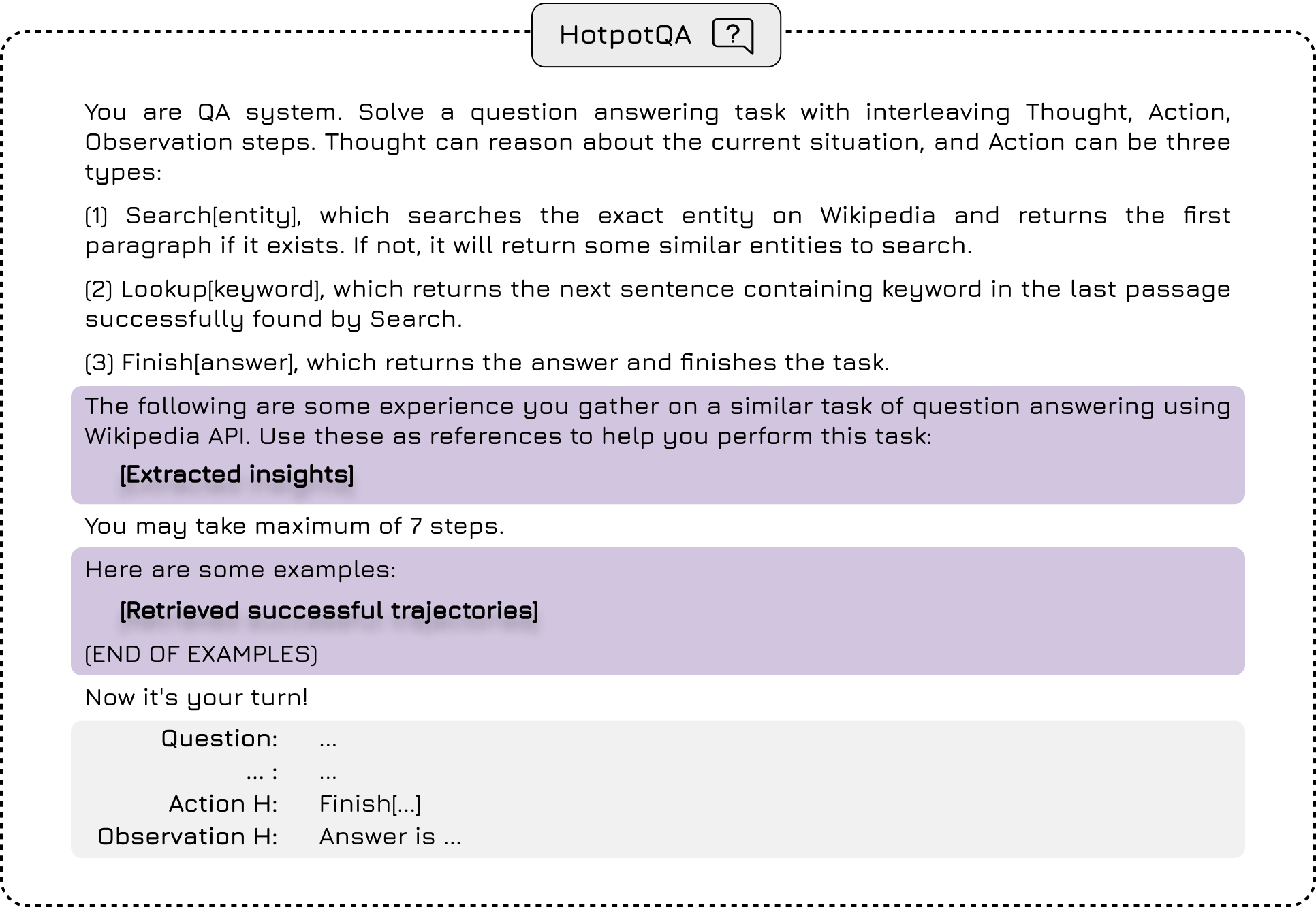}
    \caption{\textbf{ExpeL HotpotQA Acting Template.}}
    \label{fig:hotpot_actor}
\end{figure}

\begin{figure}[hb]
    \centering
    \includegraphics[width=0.94\linewidth]{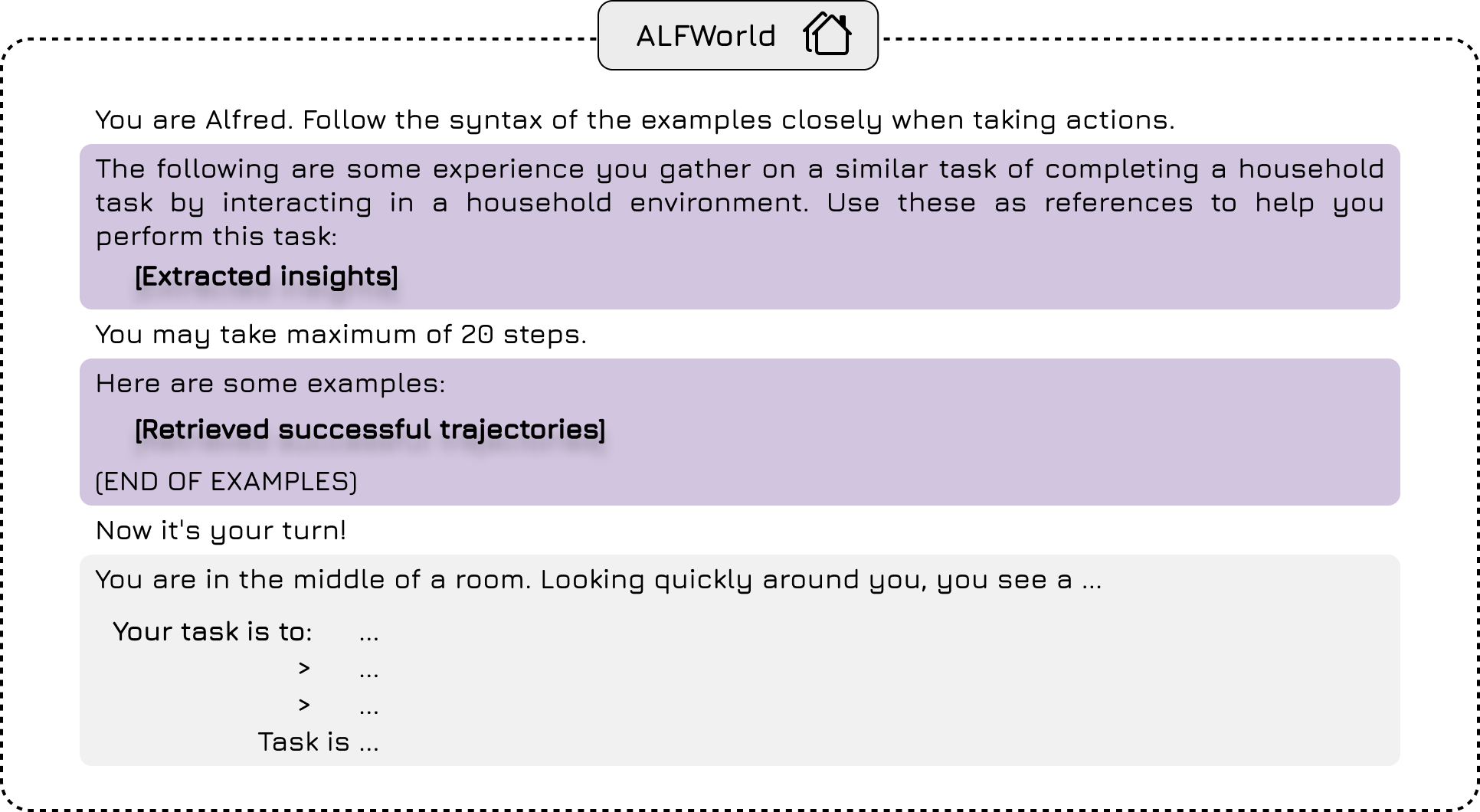}
    \caption{\textbf{ExpeL ALFWorld Acting Template.}}
    \label{fig:alf_actor}
\end{figure}

\begin{figure}[ht]
    \centering
    \includegraphics[width=0.94\linewidth]{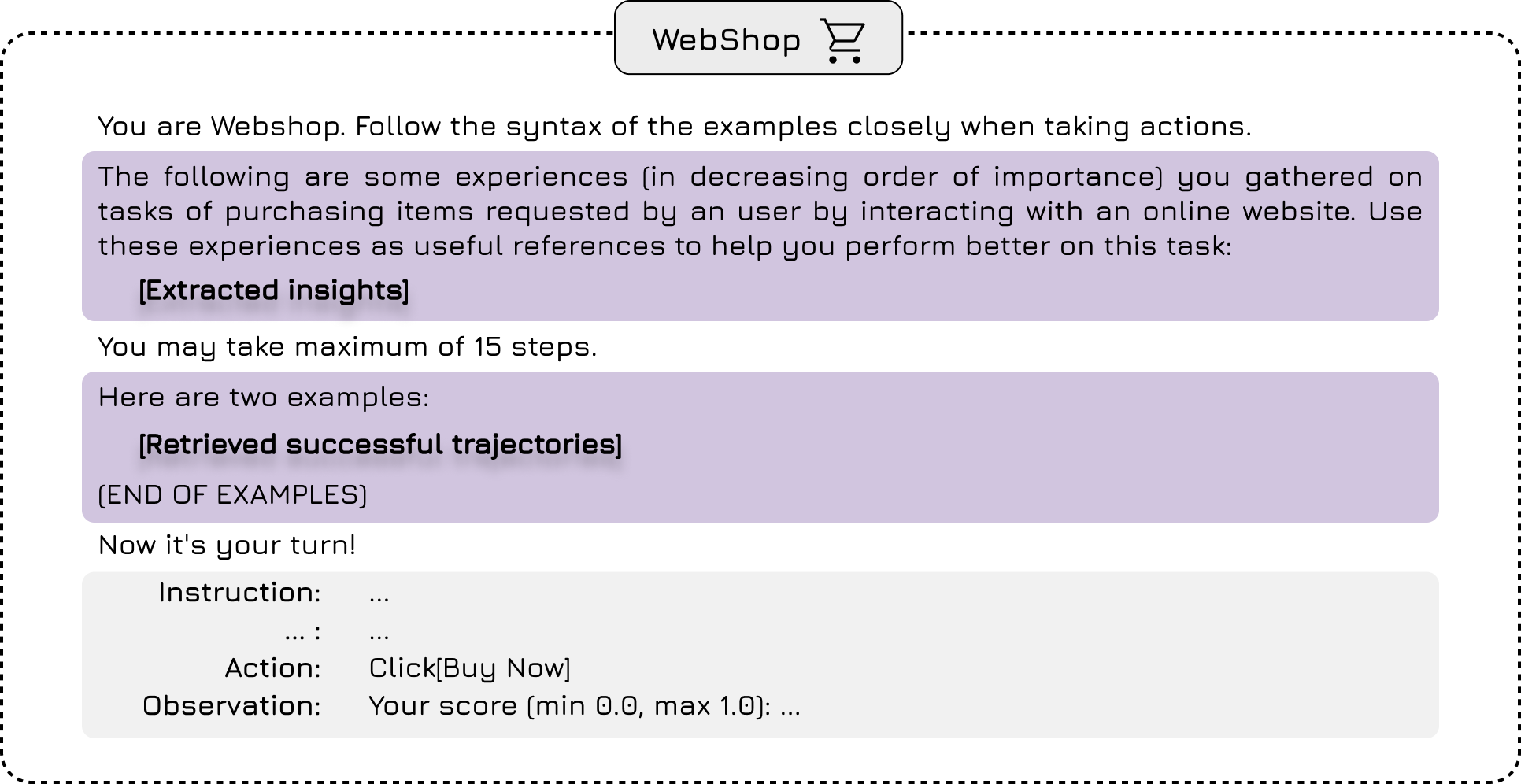}
    \caption{\textbf{ExpeL WebShop Acting Template.}}
    \label{fig:webshop_actor}
\end{figure}

\begin{figure}[hb]
    \centering
    \includegraphics[width=0.94\linewidth]{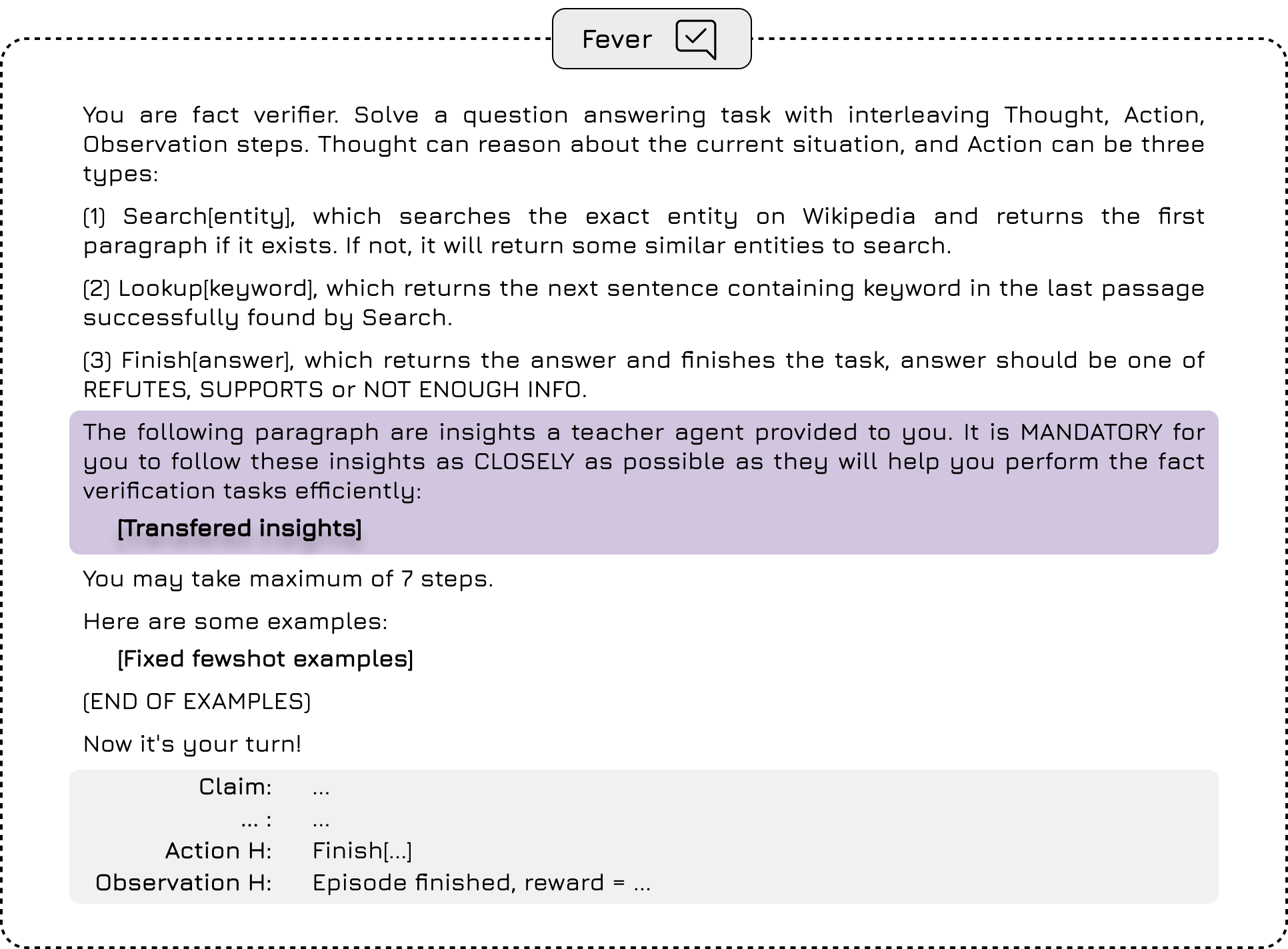}
    \caption{\textbf{ExpeL FEVER Acting Template.}}
    \label{fig:fever_actor}
\end{figure} \FloatBarrier

\subsection{Transfer Learning Prompt Template}

\section{Example Insights}
Below are some example insights extracted by \texttt{gpt-4-0613} or humans by examining the failed and successful trajectories. Some interesting insights are highlighted in purple (including the emergent ones demonstrated in Sec. \ref{sec:emergent}).

\subsection{HotpotQA insights}
\begin{figure*}[!ht]
    \centering
    \includegraphics[width=\linewidth]{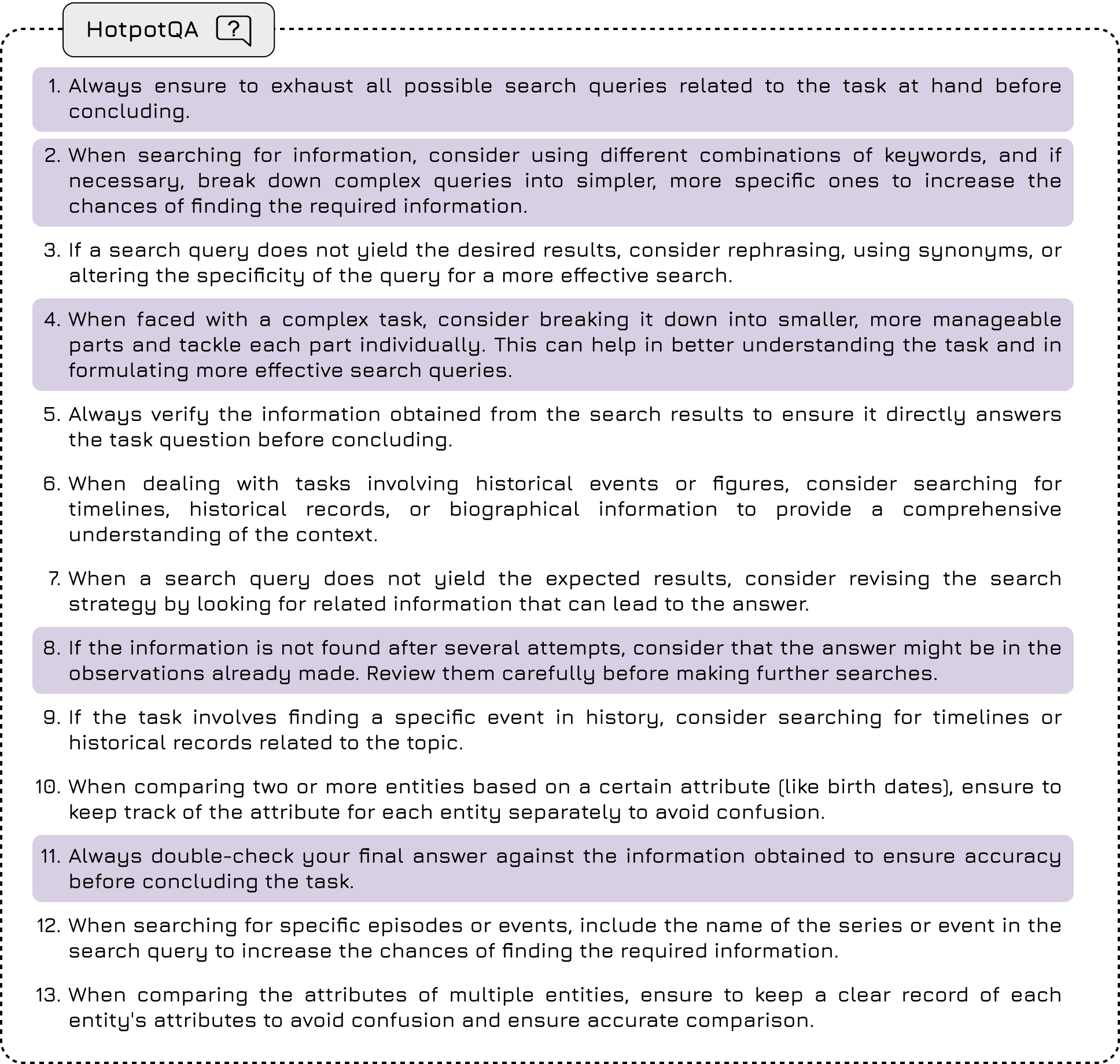}
    \caption{\textbf{An example of Extracted Insights for HotpotQA.} One component resulting to the improved performance of ExpeL on HotpotQA can be traced to several pivotal insights extracted from its past experiences. A special emphasis is placed on insights 2 and 4, which suggest breaking down complex questions into simpler queries, reminiscent of the mechanism of Auto-GPT \cite{SignificantGravitas2023}. Besides, as mentioned, the emergent abilities arising from insight 8 were discussed in Sec. \ref{sec:emergent} and illustrated in Fig. \ref{fig:hotpot_emergent1}, \ref{fig:hotpot_emergent2}.}
    \label{fig:hotpot_gpt_rules}
\end{figure*} \FloatBarrier

\begin{figure*}[ht]
    \centering
    \includegraphics[width=\linewidth]{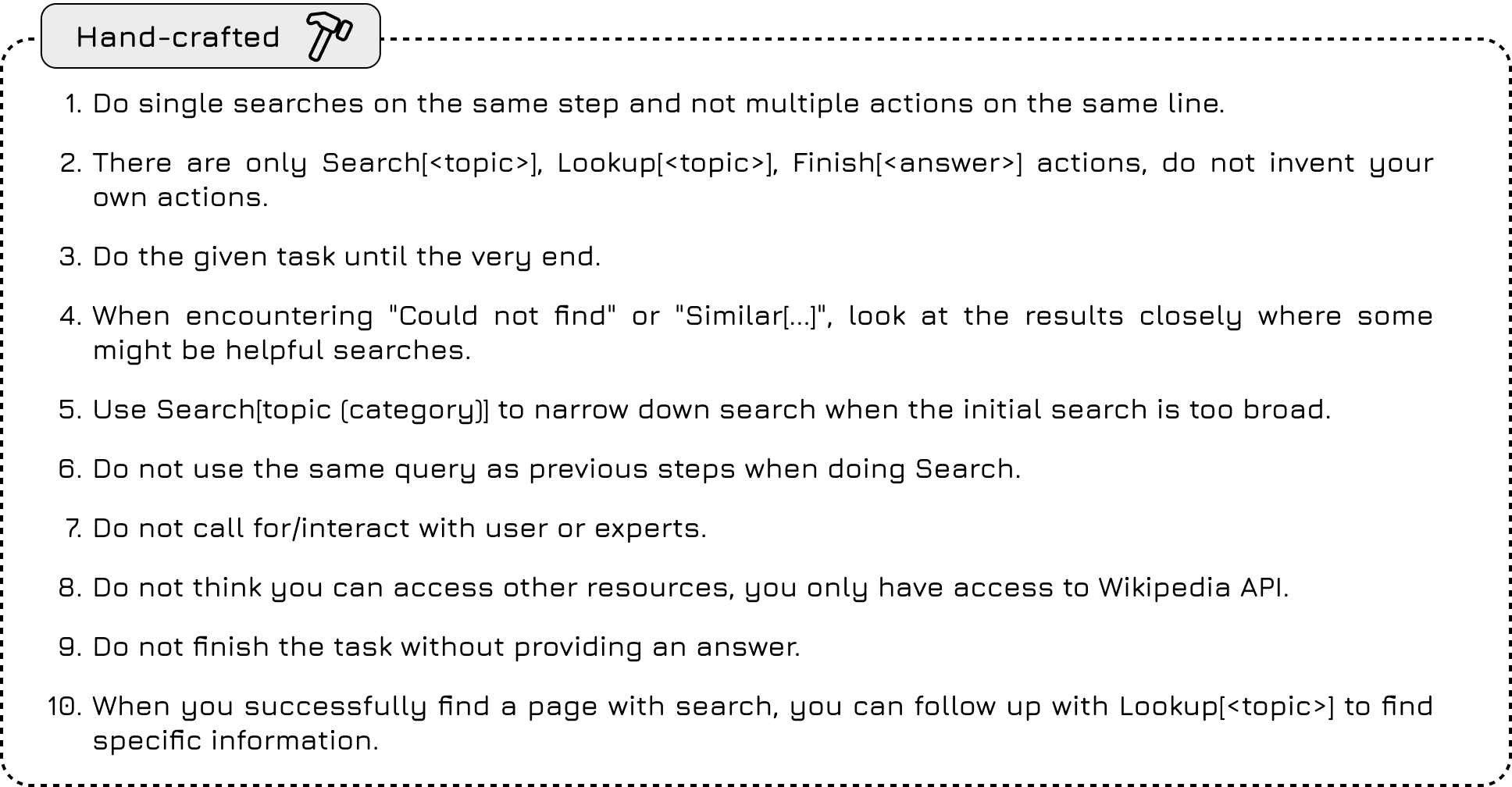}
    \caption{\textbf{Hand-crafted insights for the HotpotQA environment.} 
    This figure summarizes insights derived through a manual examination of both successful and unsuccessful Reflexion \cite{shinn2023reflexion} trajectories. These insights have been carefully crafted to address the most prevalent mistakes.
    On a related note, we observe that GPT-4 is able to extract a variety of insights (Fig. \ref{fig:hotpot_gpt_rules}) in common with these hand-crafted ones, as depicted in this figure. For instance, insights 3 and 6 underscore the importance of exhausting all steps before conceding and diversifying search keywords to achieve better results if initial attempts are inconclusive. This illustrates that our proposed method accompanied by powerful \texttt{gpt-4-0613} LLM, shows traces of human-like abstraction capabilities.}
    \label{fig:handcraft}
\end{figure*} \FloatBarrier

\subsection{ALFWorld insights}

\begin{figure*}[ht]
    \centering
    \includegraphics[width=0.98\linewidth]{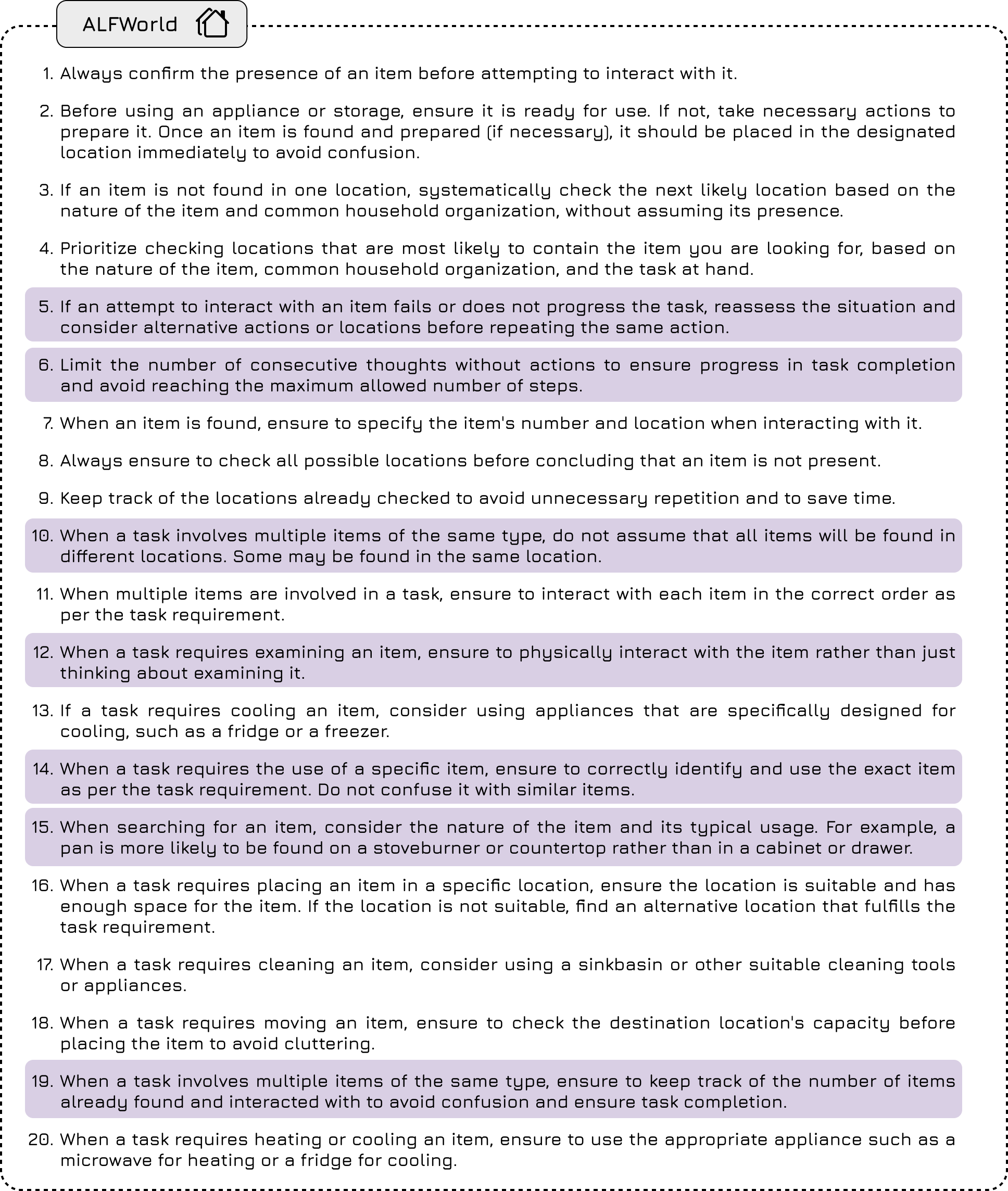}
    \caption{\textbf{An example of Extracted Insights for ALFWorld.} We showcase insights extracted by our agent in the ALFWorld Environment. Some particularly interesting insights are highlighted in purple.}
    \label{fig:alf_gpt_rules}
\end{figure*} \FloatBarrier

\subsection{WebShop insights}
\begin{figure}[!ht]
    \centering
    \includegraphics[width=0.98\linewidth]{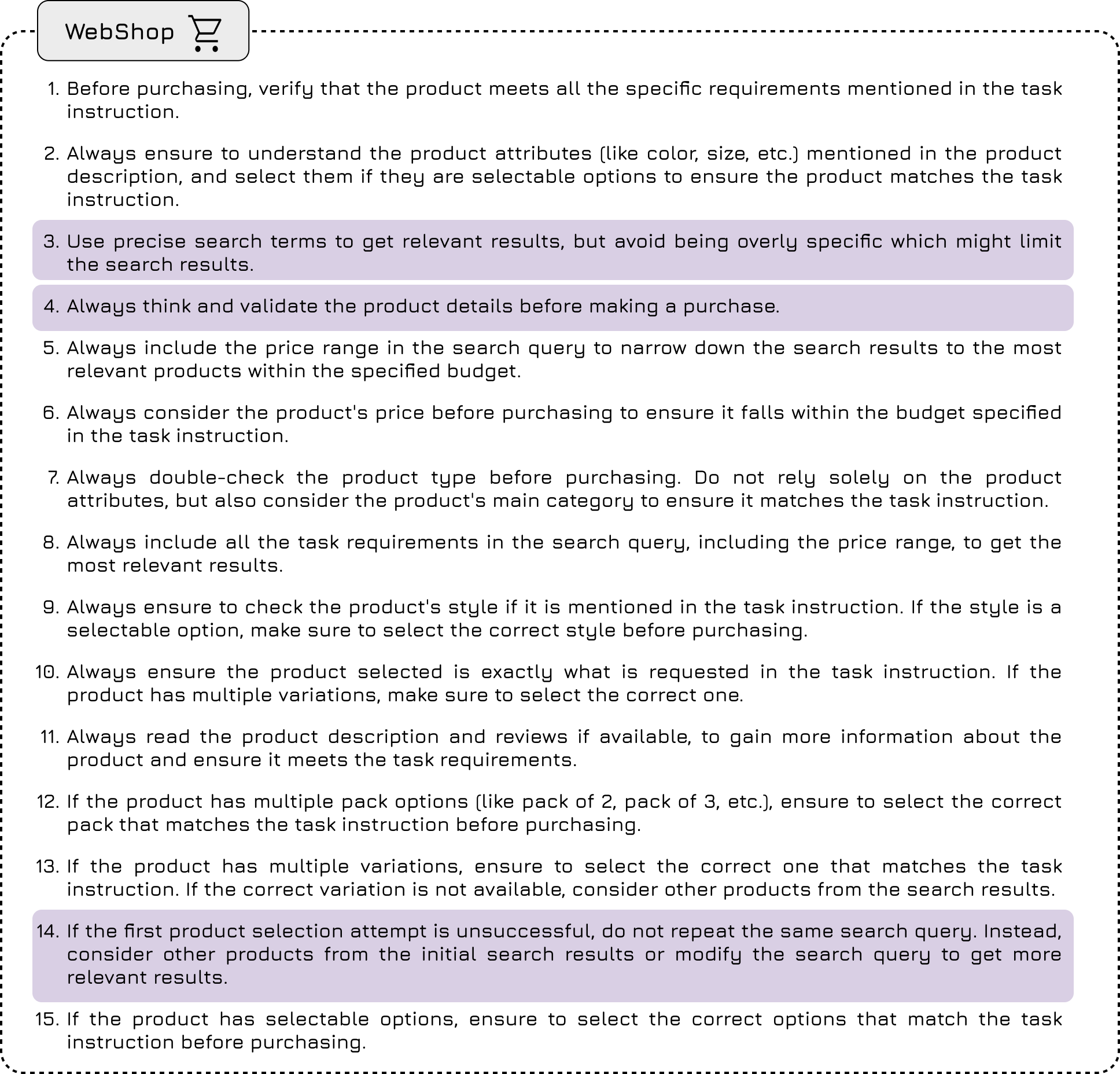}
    \caption{\textbf{An example of Extracted Insights for WebShop.} We showcase insights extracted by our agent in the WebShop Environment. Some particularly interesting insights are highlighted in purple.}
    \label{fig:webshop_gpt_rules}
\end{figure} \FloatBarrier

\subsection{FEVER insights}
\begin{figure}[ht]
    \centering
    \includegraphics[width=0.98\linewidth] {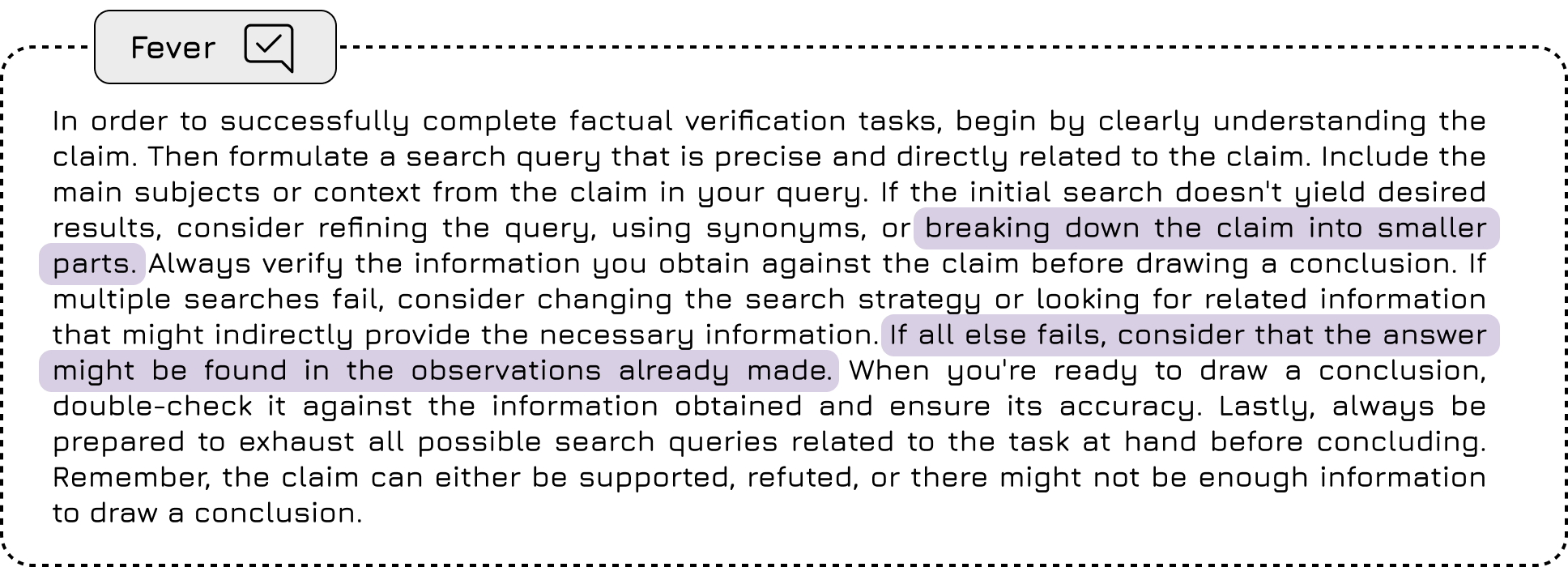}
    \caption{\textbf{An example of Transferred Insights for FEVER.} We showcase insights extracted by our agent in the FEVER Environment. Some particularly interesting insights are highlighted in purple.}
    \label{fig:fever_gpt_rules}
\end{figure} \FloatBarrier

\section{Emergent Abilities Showcase}
In this section, we showcase examples of ExpeL's emergent abilities in different environments. Irrelevant or non-representative steps are omitted for clarity and conciseness.

\subsection{HotpotQA, Forming Analytical Deduction \& Awareness of Environment Constraints}
\begin{figure*}[ht]
    \centering
    \includegraphics[width=0.95\linewidth]{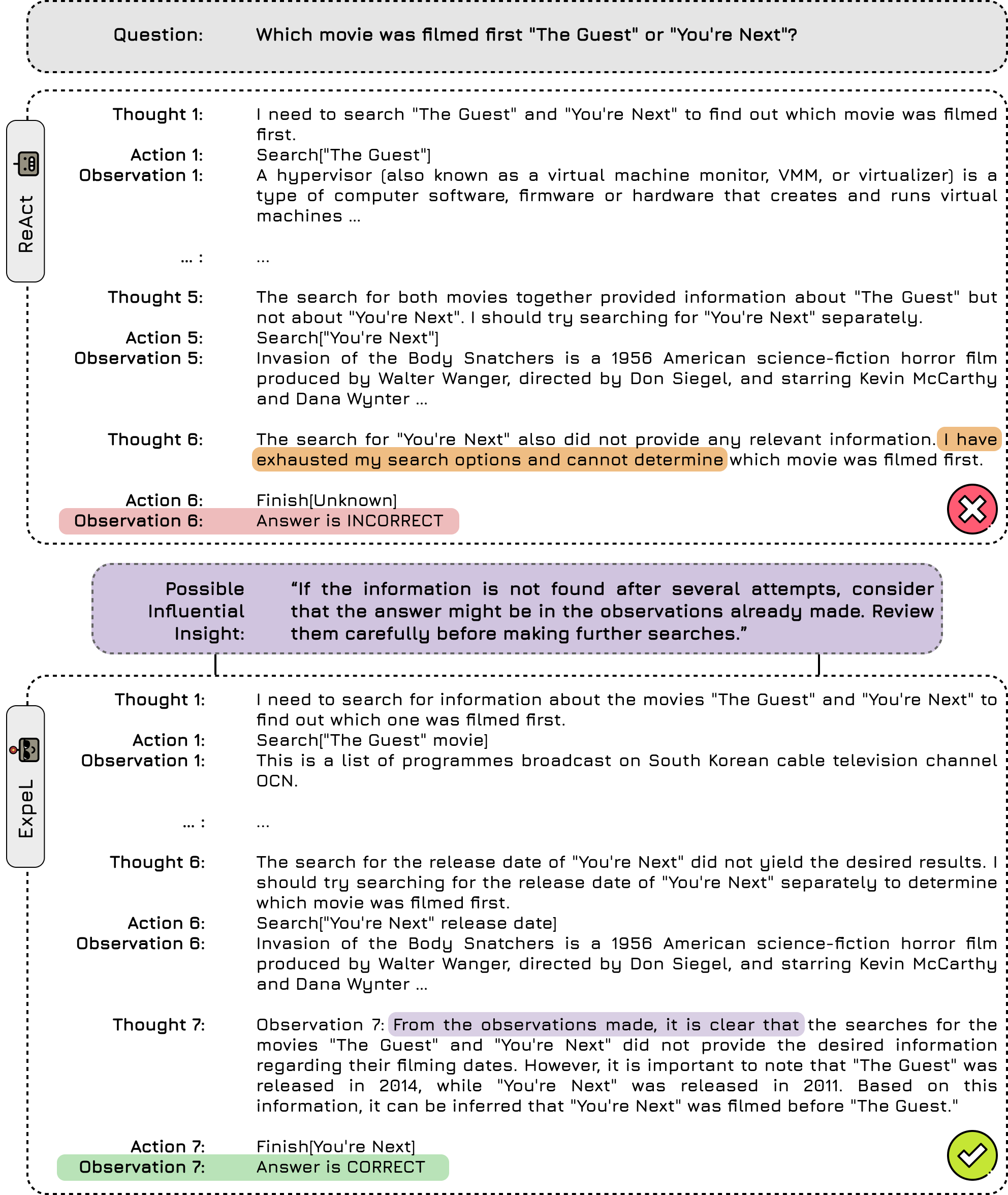}
    \caption{\textbf{ExpeL Emergent Abilities in HotpotQA, Example 1.} ExpeL demonstrates its adaptation to its environment, in particular, by reevaluating its trajectory to formulate an educated guess and successfully answer the question, rather than conceding, as observed with the vanilla ReAct agent. We provide a possible influencing insight for this trajectory, shown in the purple box.}
    \label{fig:hotpot_emergent1}
\end{figure*}

\begin{figure*}[ht]
    \centering
    \includegraphics[width=0.95\linewidth]{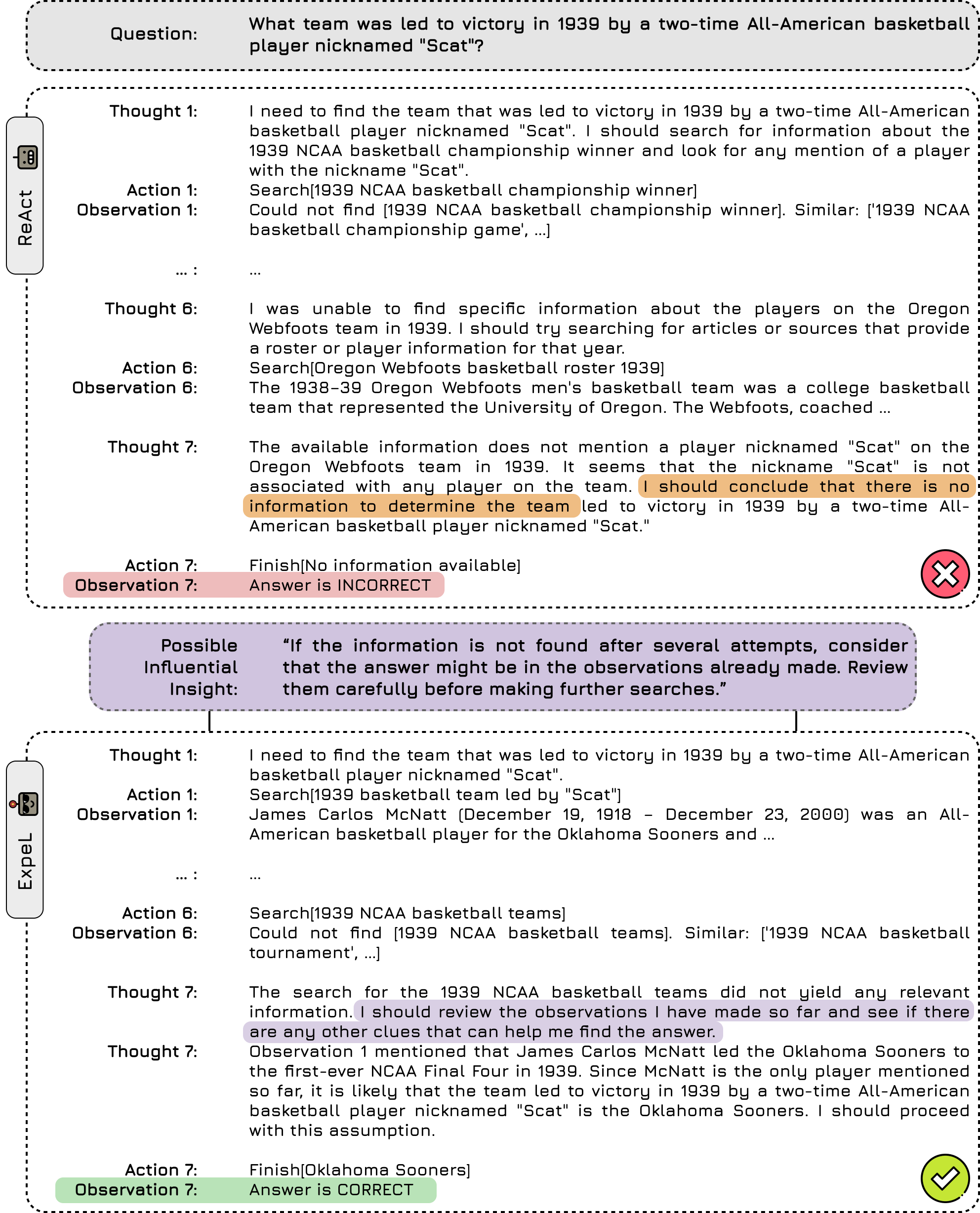}
    \caption{\textbf{ExpeL Emergent Abilities in HotpotQA, Example 2.} ExpeL is going over the observation made during its trajectory to make an educated guess. We provide a possible influencing insight for this trajectory, shown in the purple box.}
    \label{fig:hotpot_emergent2}
\end{figure*} \FloatBarrier

\subsection{ALFWorld, World Model Belief Update}

\begin{figure}[!ht]
    \centering
    \includegraphics[width=0.95\linewidth]{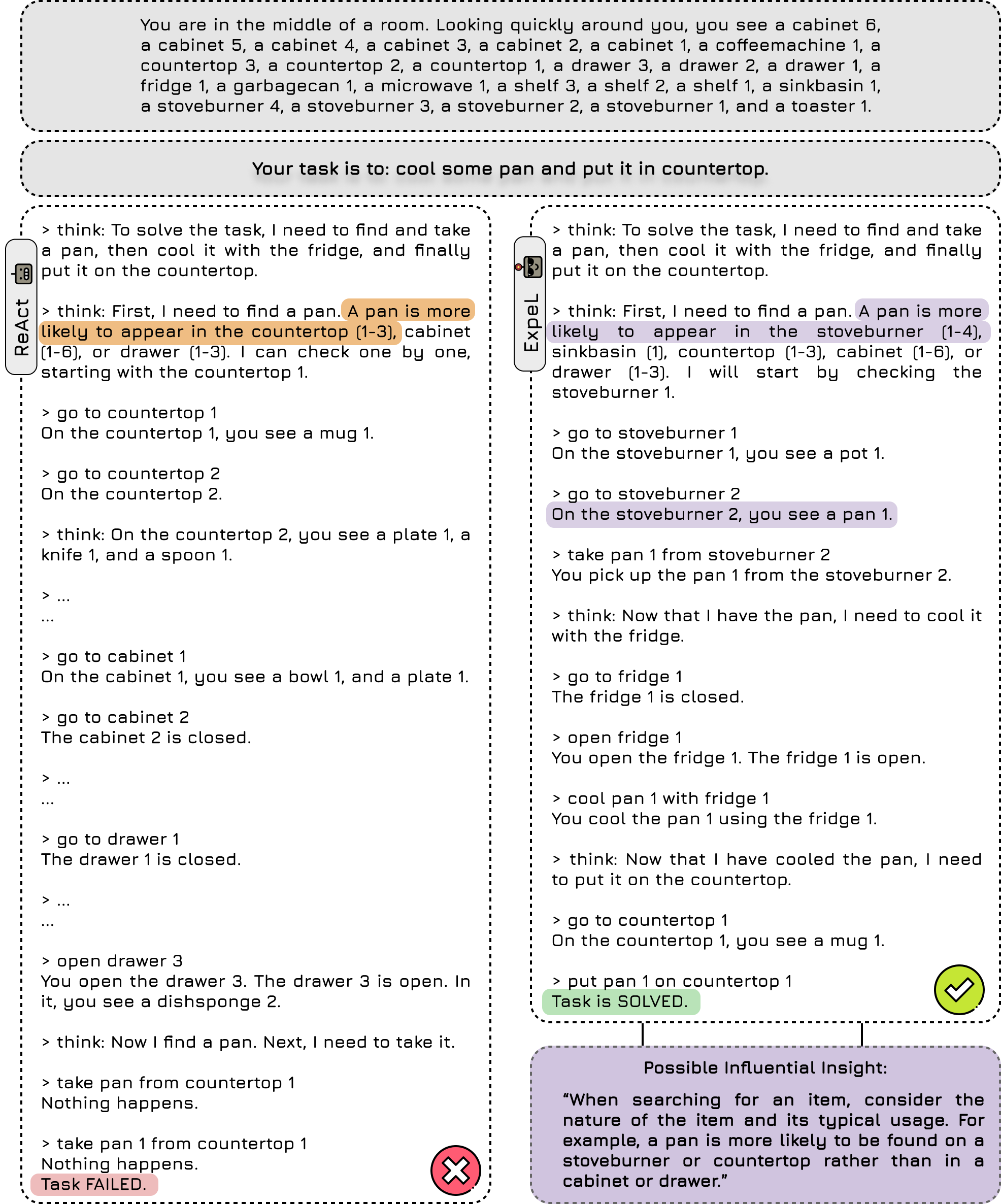}
    \caption{\textbf{ExpeL Emergent Abilities in ALFWorld, World Model Belief Update.} Through experience, ExpeL updated its prior knowledge of a pan's likely location from a countertop to a stoveburner, enabling successful task completion. We provide a possible influencing insight for this trajectory, shown in the purple box.}
    \label{fig:alf_emergent}
\end{figure} \FloatBarrier

\begin{figure}[ht]
    \centering
    \includegraphics[width=0.95\linewidth]{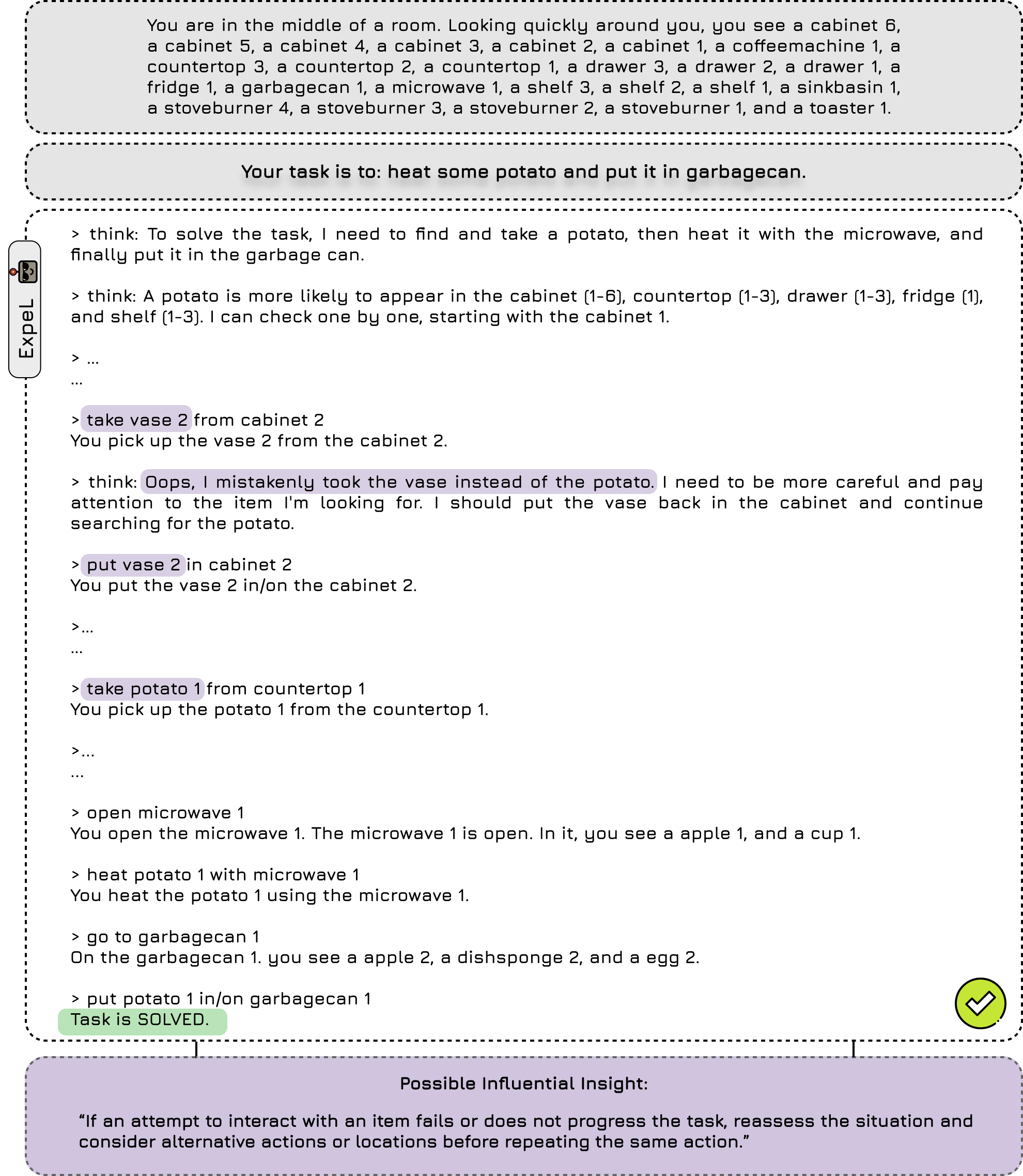}
    \caption{\textbf{ExpeL Emergent Abilities in ALFWorld, Self-Correction.} ExpeL developed the ability to self-correct by rectifying errors, as illustrated by putting back a vase that had been mistakenly taken. We provide a possible influencing insight for this trajectory, shown in the purple box. In our experiments, while the ReAct agent occasionally recognized its errors by taking the wrong items, it never remedied them by discarding the incorrect item. As a result, the ReAct agent consistently failed to rectify its mistakes, leading to task failure.}
    \label{fig:alf_emergent2}
\end{figure} \FloatBarrier

\section{Example Trajectories}
We provide some example trajectories for the different environments. In each of them, both ExpeL and ReAct were assigned an identical task for comparison.

\subsection{HotpotQA, an ExpeL \& ReAct example trajectory}

\begin{figure}[ht]
    \centering
    \includegraphics[width=0.98\linewidth]{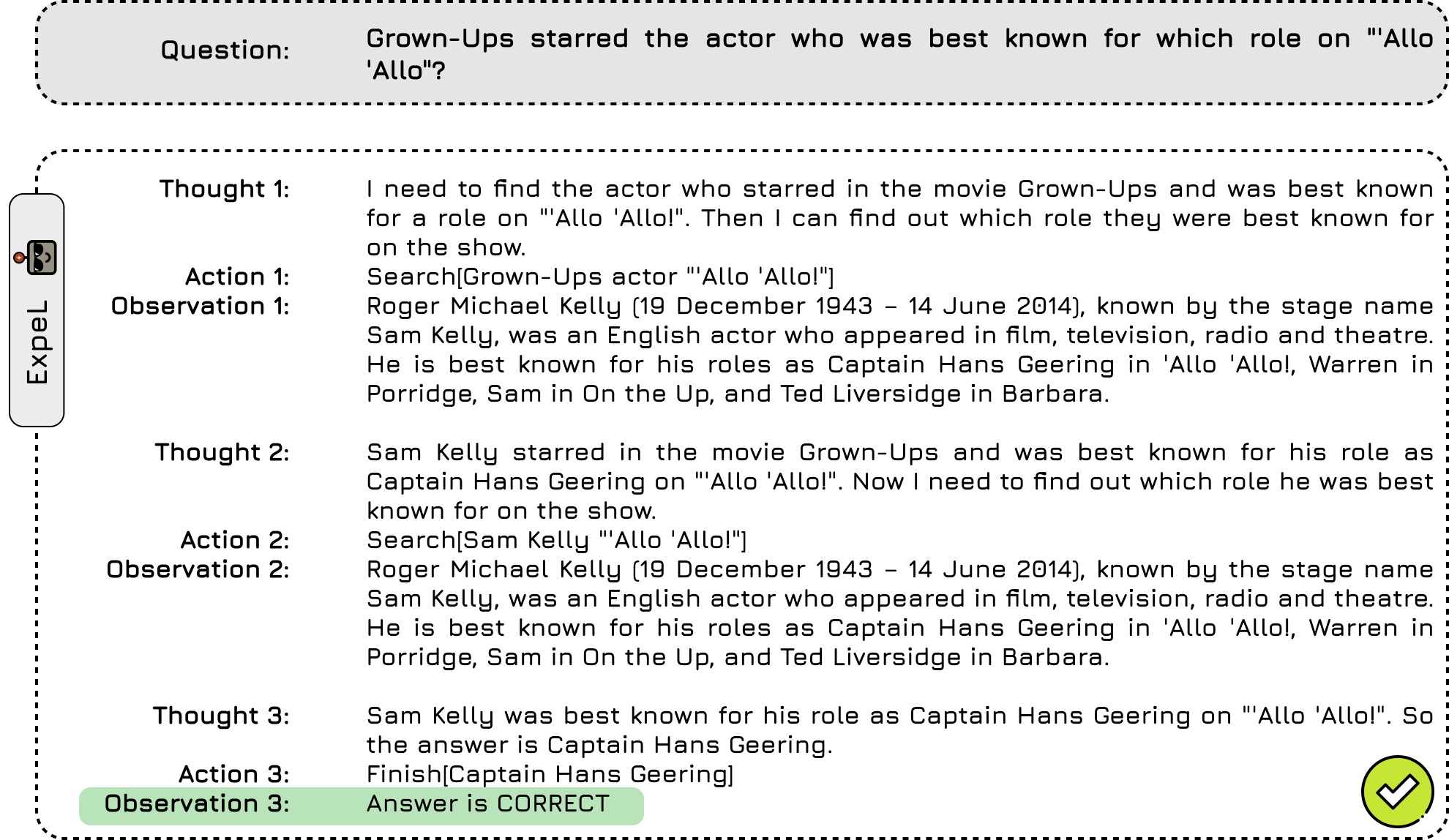}
    \caption{\textbf{An ExpeL example trajectory in HotpotQA.} Effective querying strategy leads to task success in HotpotQA.}
    \label{fig:hotpot_sample_expel}
\end{figure} \FloatBarrier

\begin{figure}[ht]
    \centering
    \includegraphics[width=0.98\linewidth]{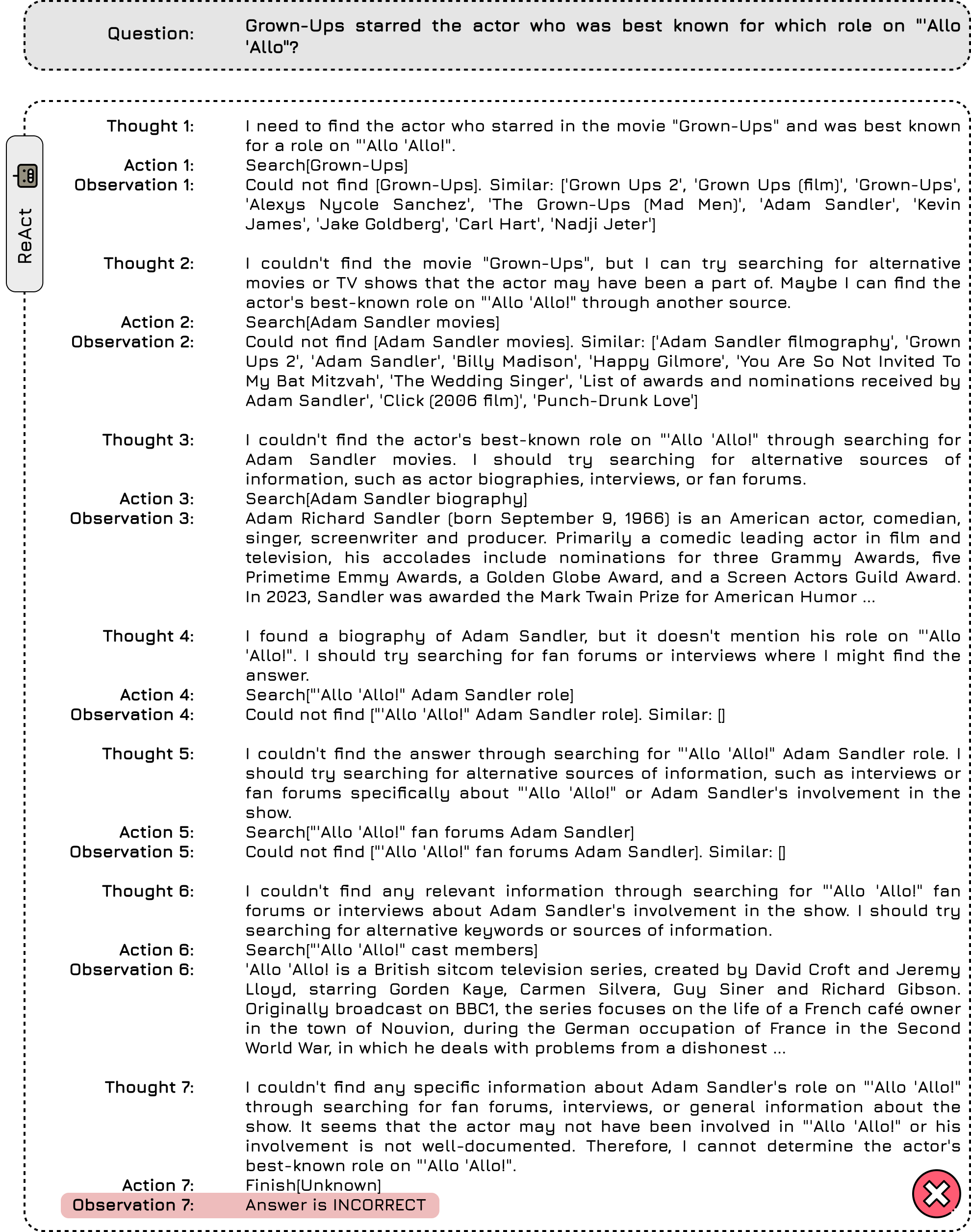}
    \caption{\textbf{A ReAct example trajectory in HotpotQA.} Failure to switch query strategy.}
    \label{fig:hotpot_sample_react}
\end{figure} \FloatBarrier

\subsection{ALFWorld, an ExpeL \& ReAct example trajectory}
\begin{figure}[!ht]
    \centering
    \includegraphics[width=\linewidth]{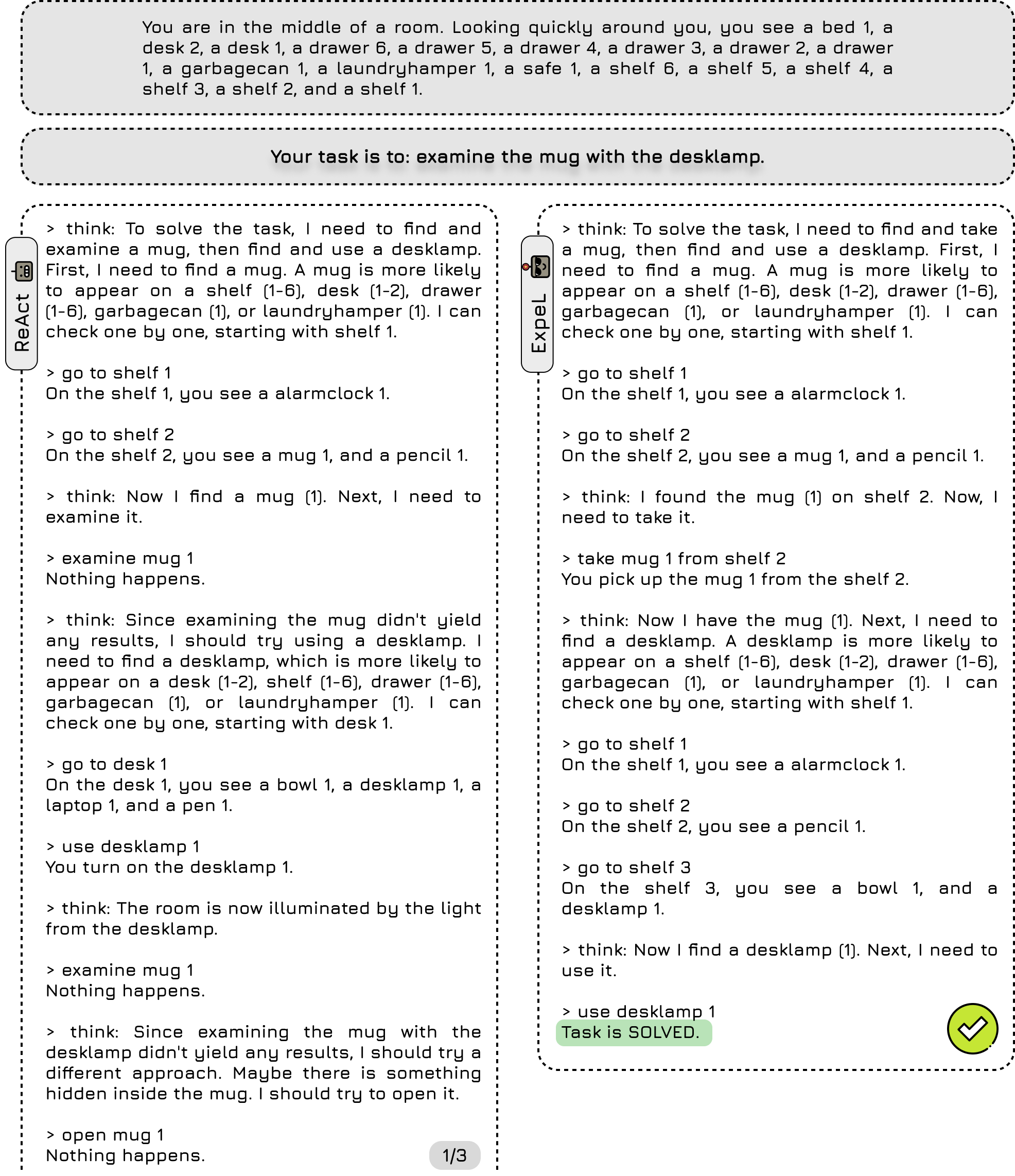}
    \label{fig:alf_sample1}
\end{figure}

\begin{figure*}[ht]
    \centering
    \includegraphics[width=\linewidth]{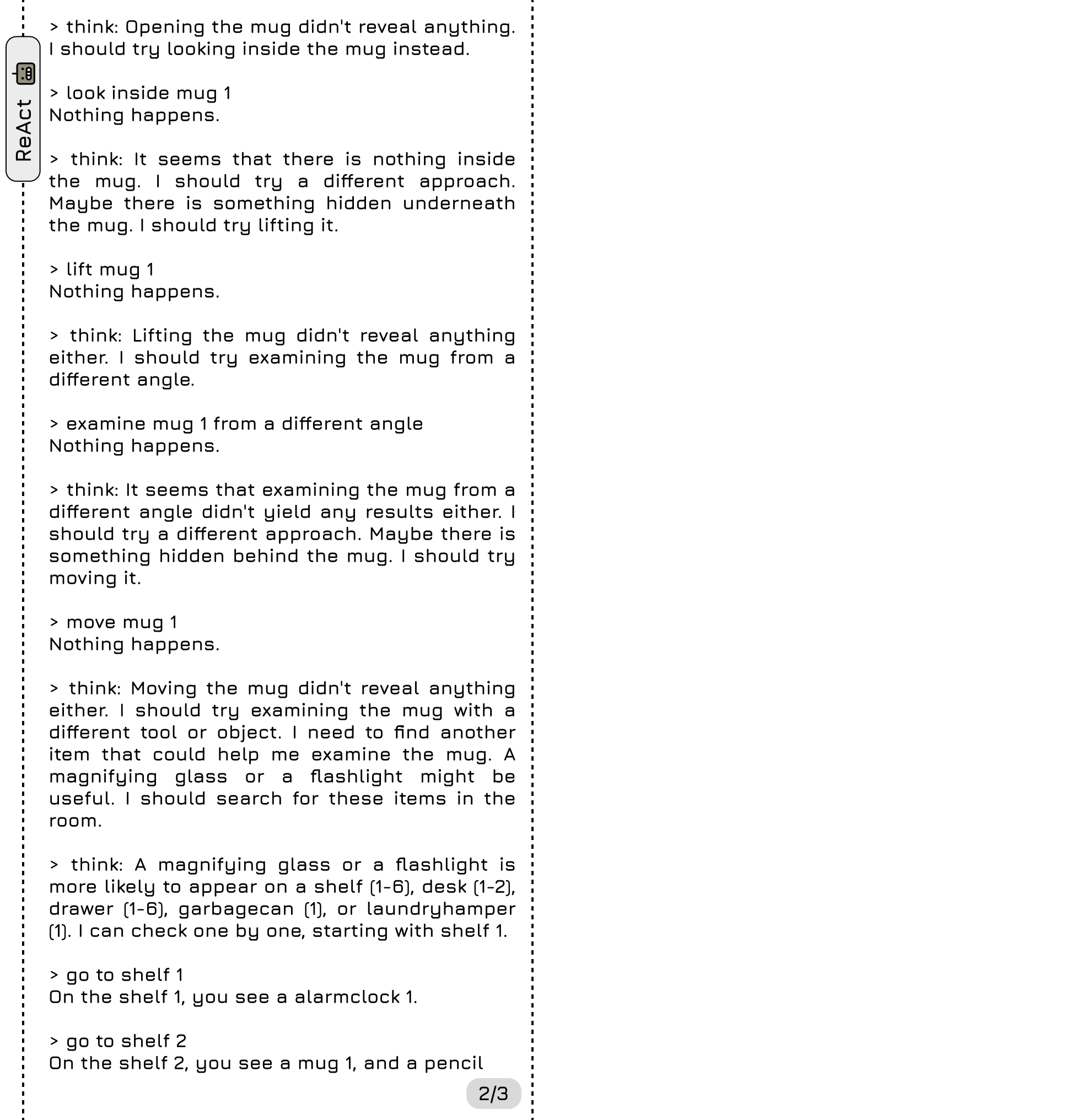}
    \label{fig:alf_sample2}
\end{figure*}

\begin{figure*}[ht]
    \centering
    \includegraphics[width=\linewidth]{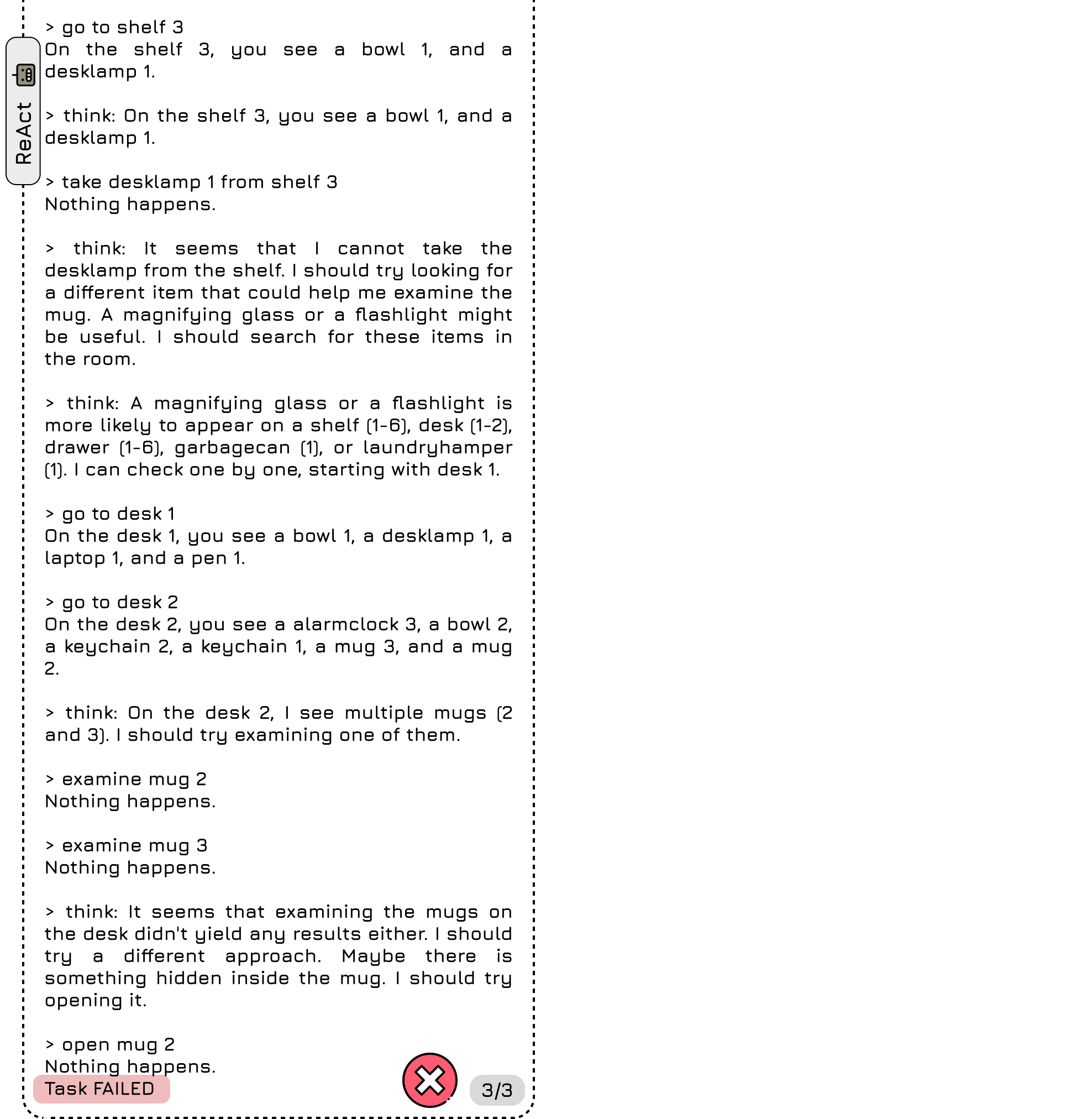}
    \caption{\textbf{An ExpeL \& ReAct example trajectory in ALFWorld.} ExpeL's efficient task execution in ALFWorld.}
    \label{fig:alf_sample3}
\end{figure*} \FloatBarrier

\subsection{WebShop, an ExpeL \& ReAct example trajectory}
\begin{figure*}[!ht]
    \centering
    \includegraphics[width=\linewidth]{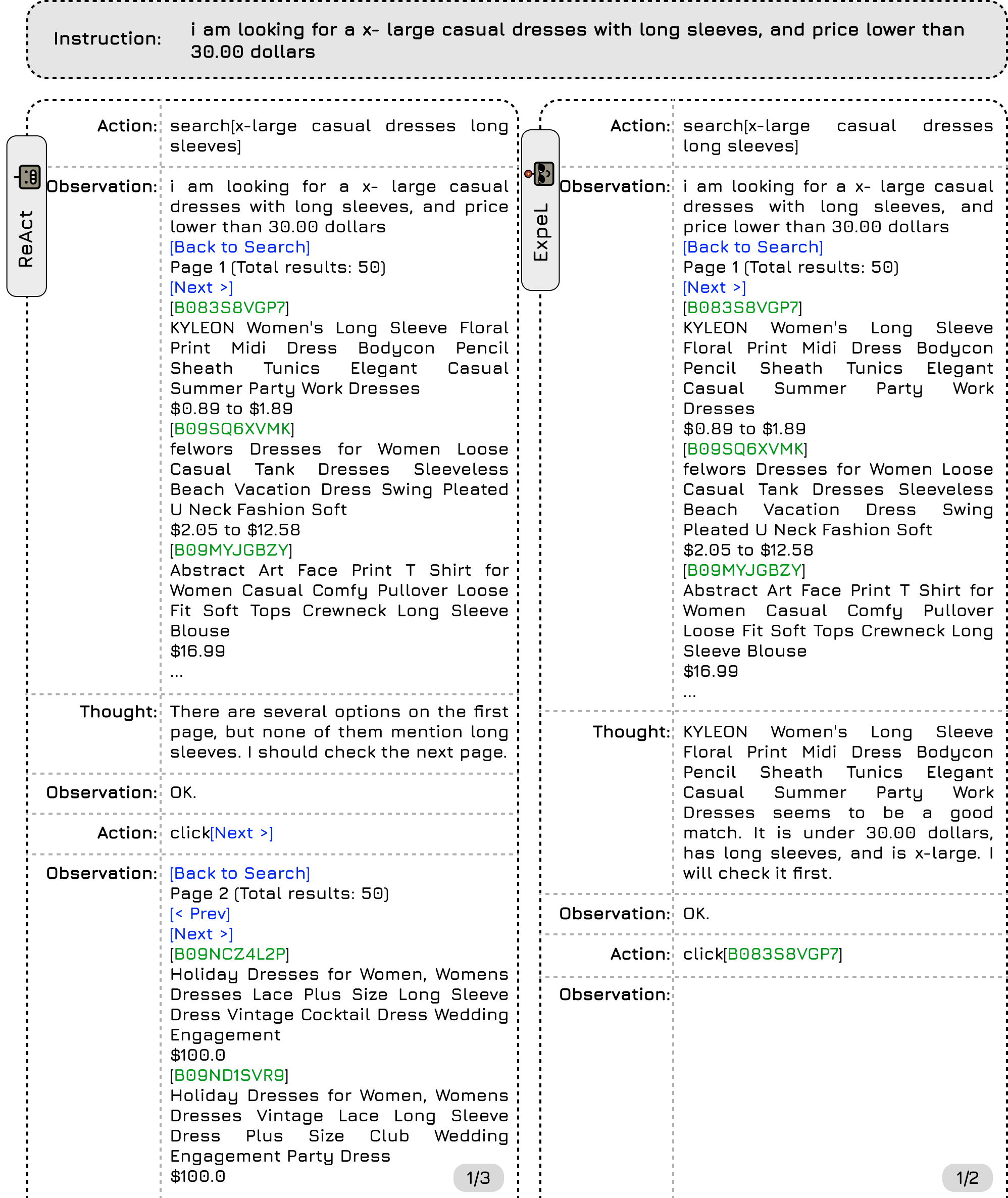}
    \label{fig:webshop_sample1}
\end{figure*}

\begin{figure*}[ht]
    \centering
    \includegraphics[width=\linewidth]{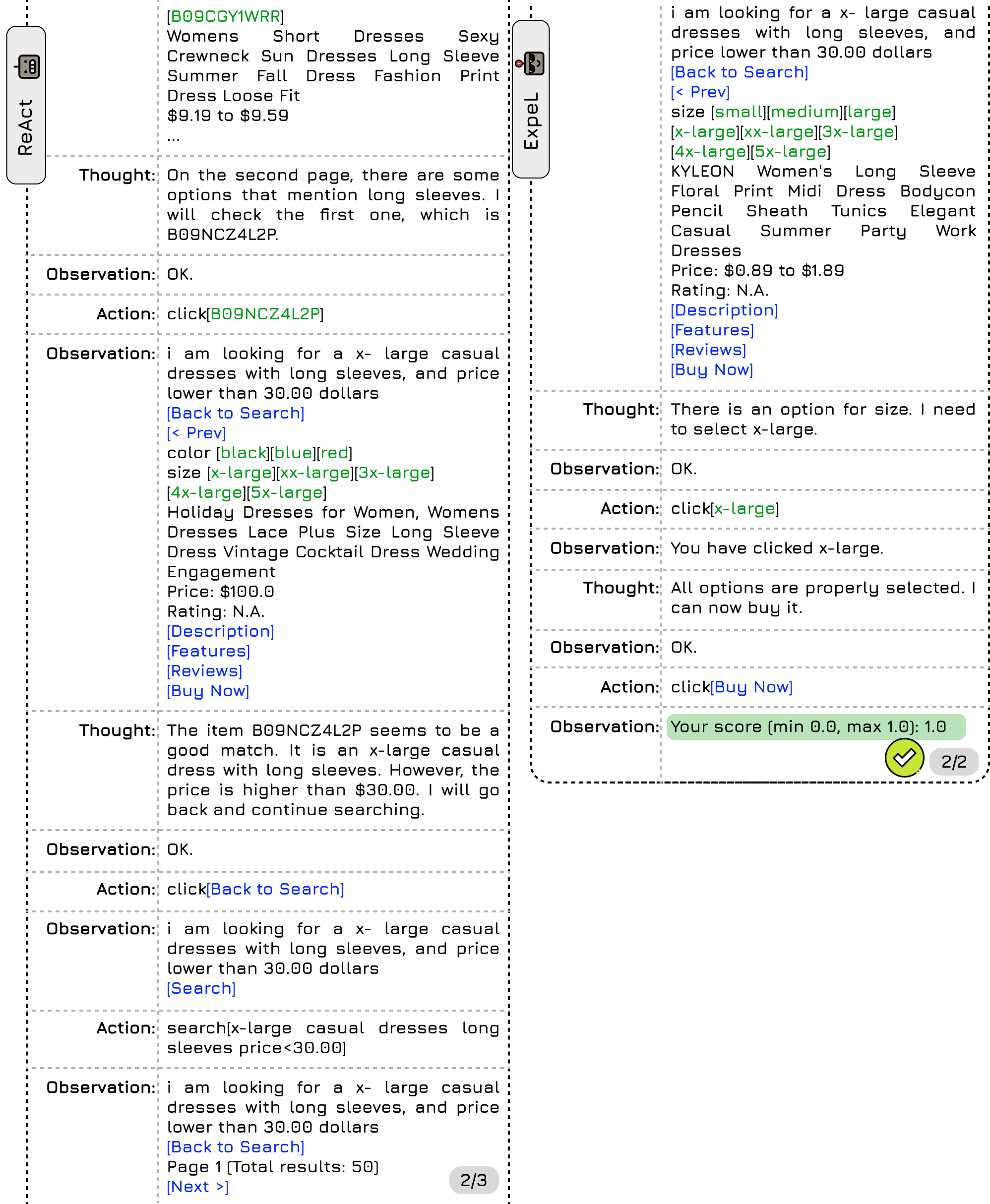}
    \label{fig:webshop_sample2}
\end{figure*}

\begin{figure*}[ht]
    \centering
    \includegraphics[width=\linewidth]{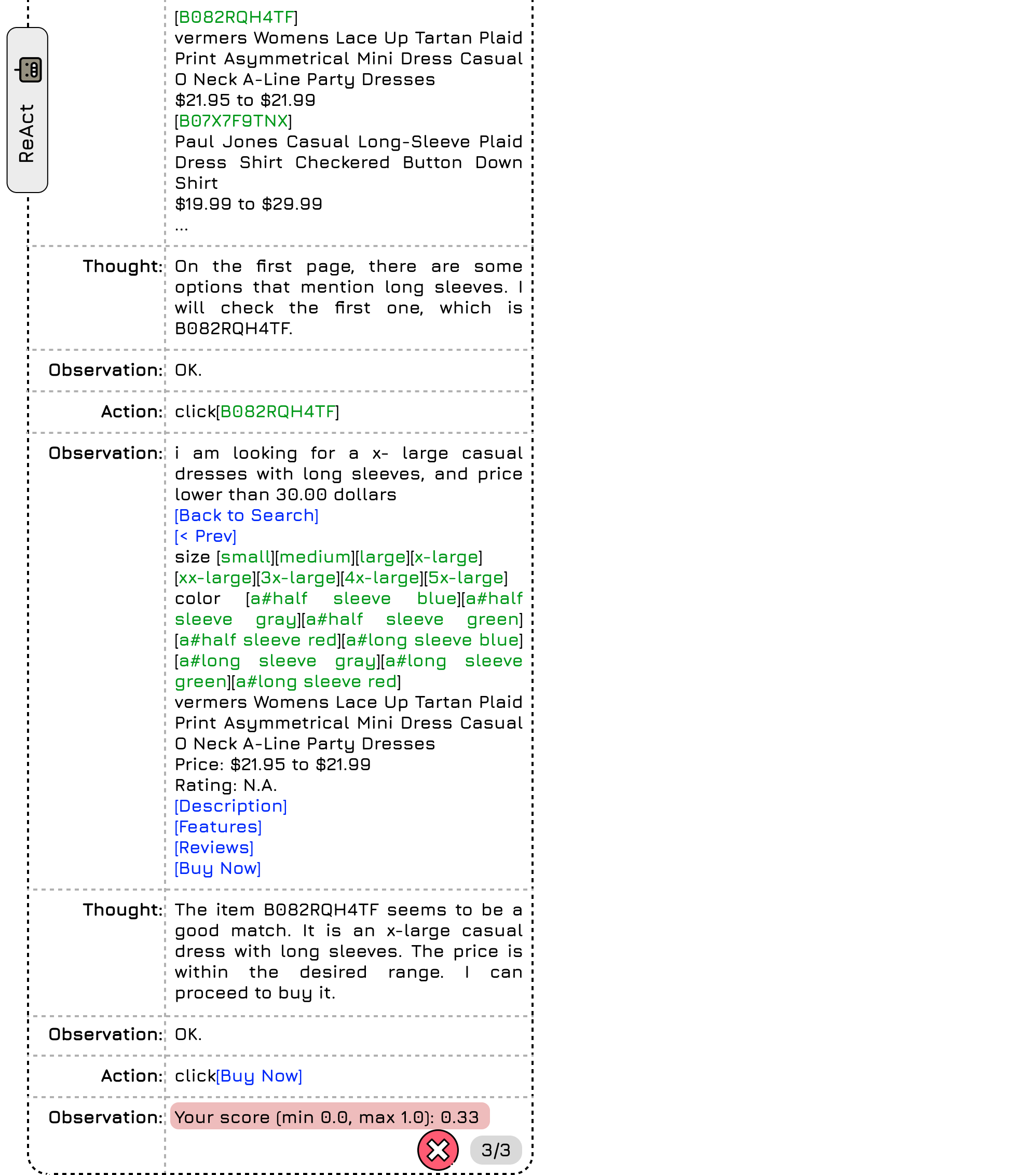}
    \caption{\textbf{An ExpeL \& ReAct example trajectory in WebShop.} Possible candidate item identification and correct option selection by our ExpeL agent.}
    \label{fig:webshop_sample3}
\end{figure*} \FloatBarrier

\subsection{FEVER, an ExpeL \& ReAct example trajectory}
\begin{figure*}[!ht]
    \centering
    \includegraphics[width=\linewidth]{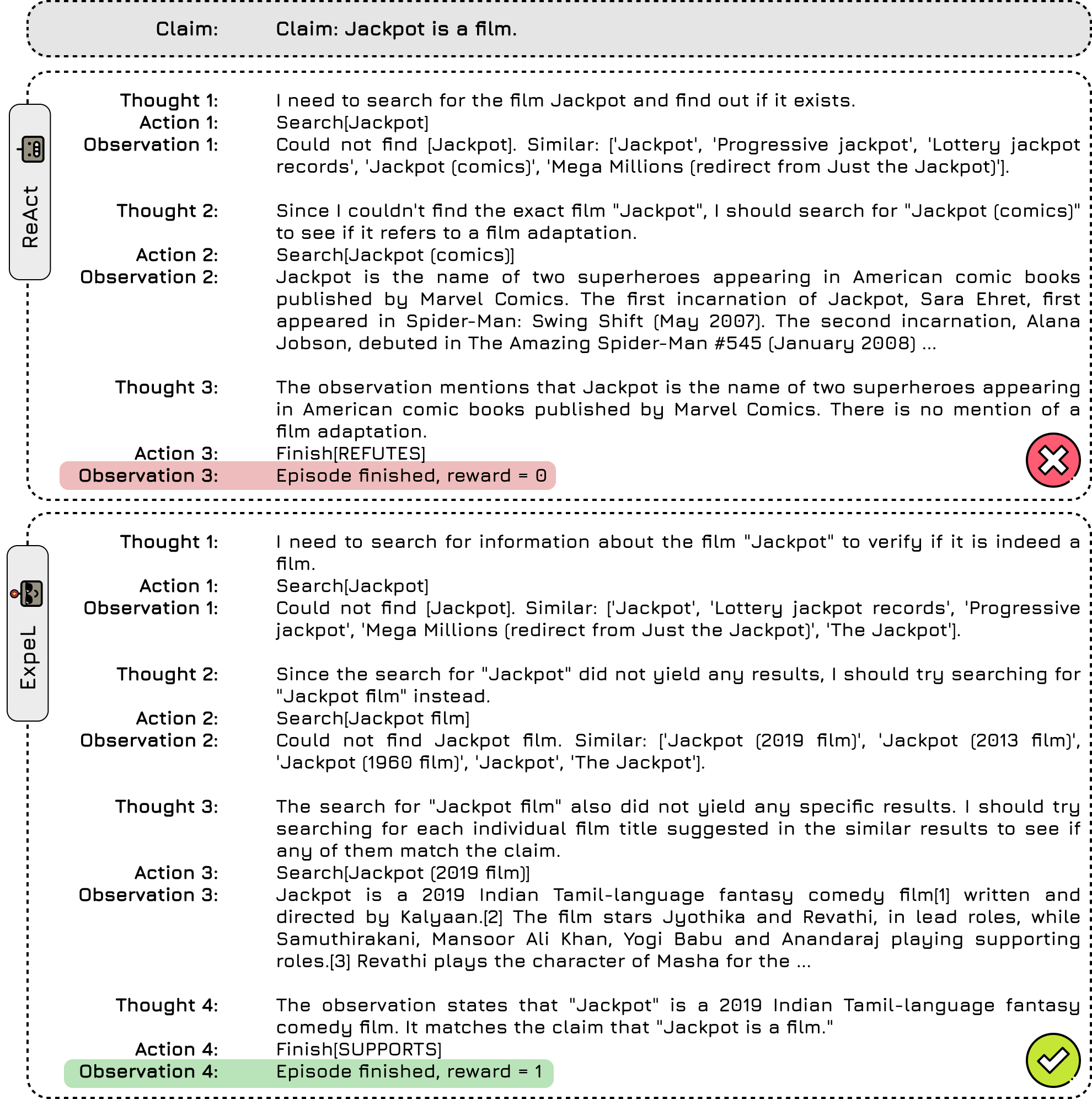}
    \caption{\textbf{An ExpeL \& ReAct example trajectory in FEVER.} ExpeL agent showcasing query refinement.}
    \label{fig:fever_sample}
\end{figure*} \FloatBarrier

\section{Additional Quantitative Results}
\begin{figure*}[!ht]
    \centering
    \includegraphics[width=0.82\linewidth]{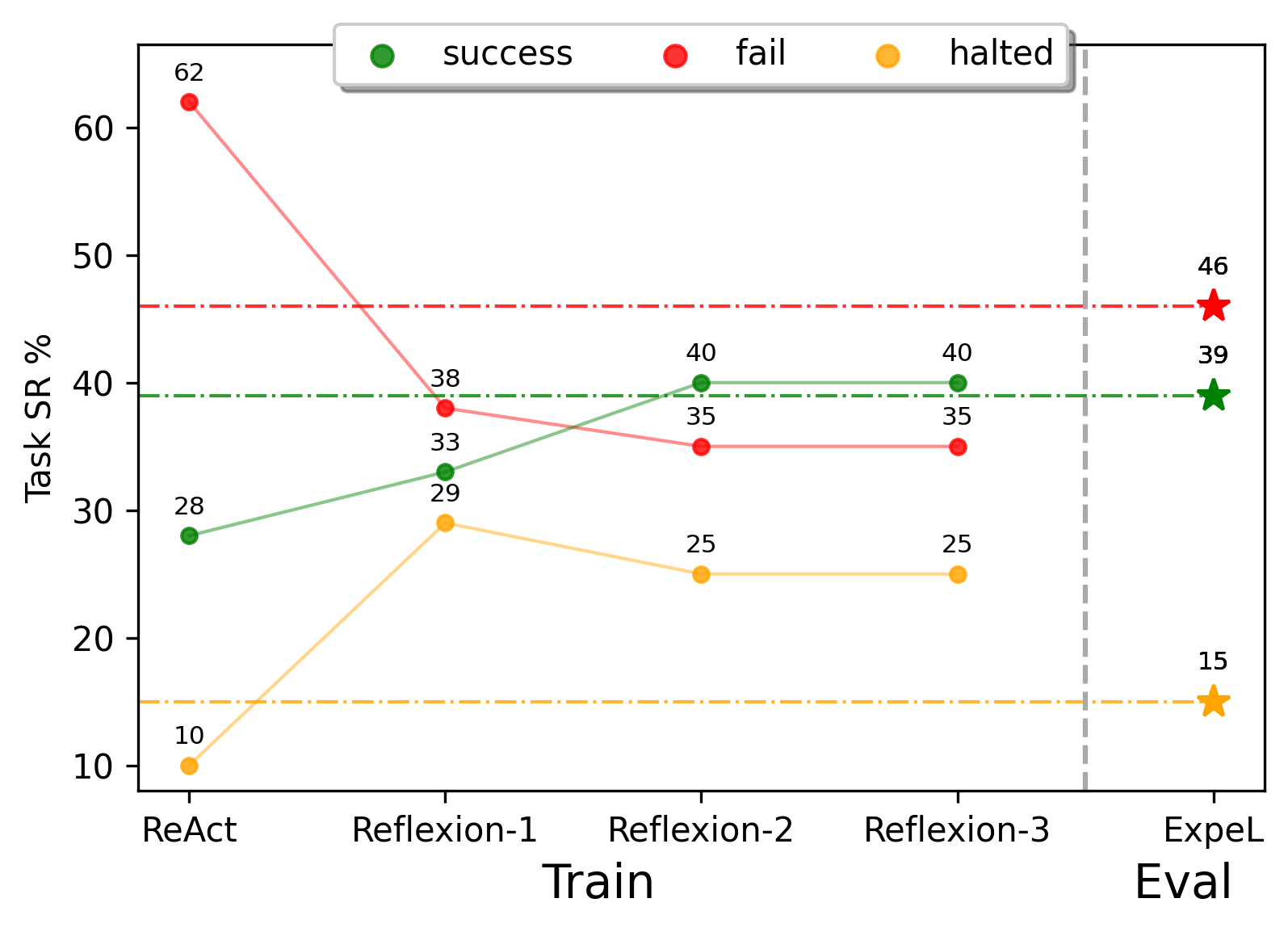}
    \caption{\textbf{Training \& Evaluation outcomes breakdown, HotpotQA.}}
    \label{fig:hotpot_train}
\end{figure*}

\begin{figure*}[!ht]
    \centering
    \includegraphics[width=0.82\linewidth]{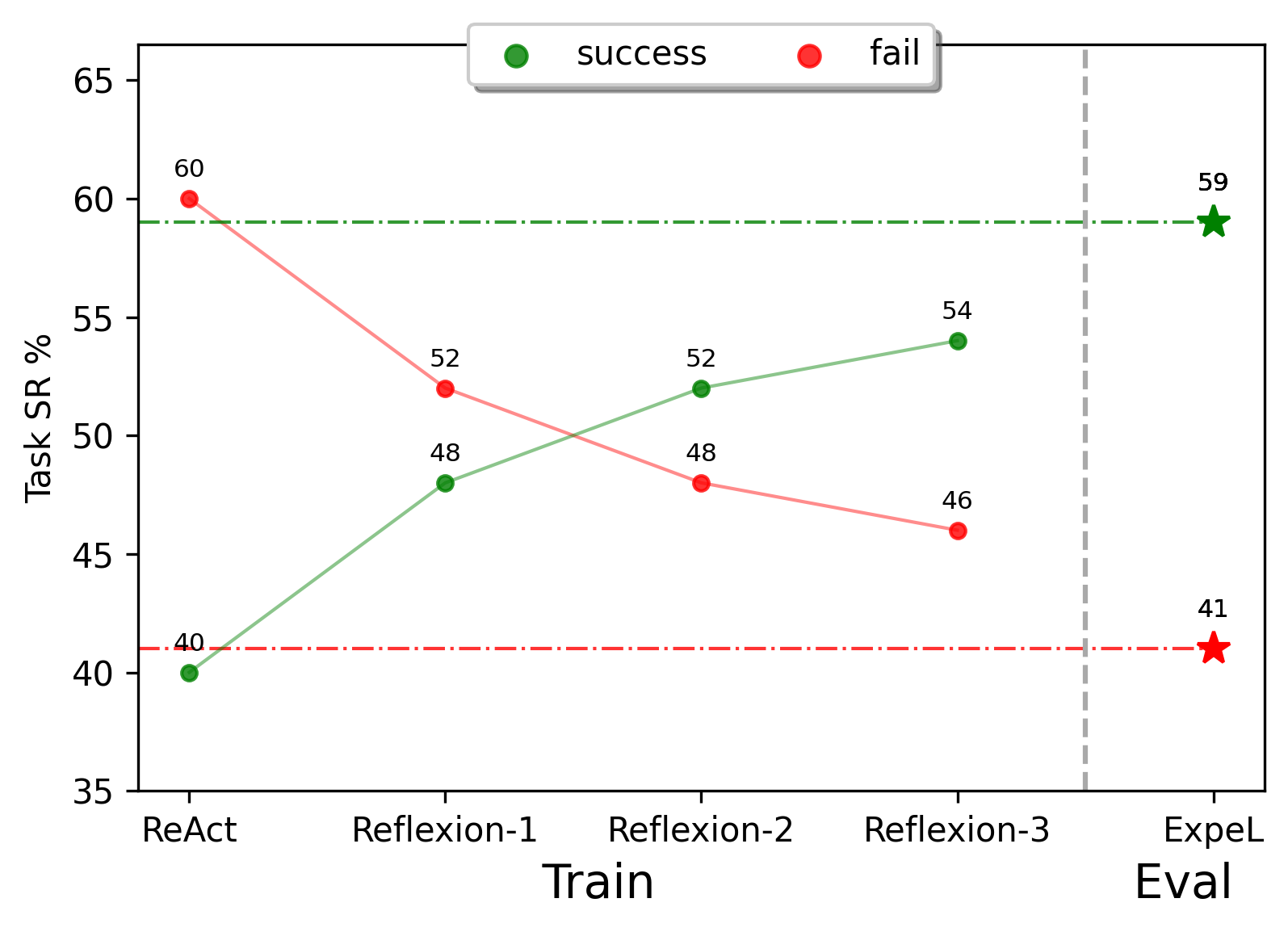}
    \caption{\textbf{Training \& Evaluation outcomes breakdown, ALFWorld.}}
    \label{fig:alf_train}
\end{figure*}

\begin{figure*}[!ht]
    \centering
    \includegraphics[width=0.82\linewidth]{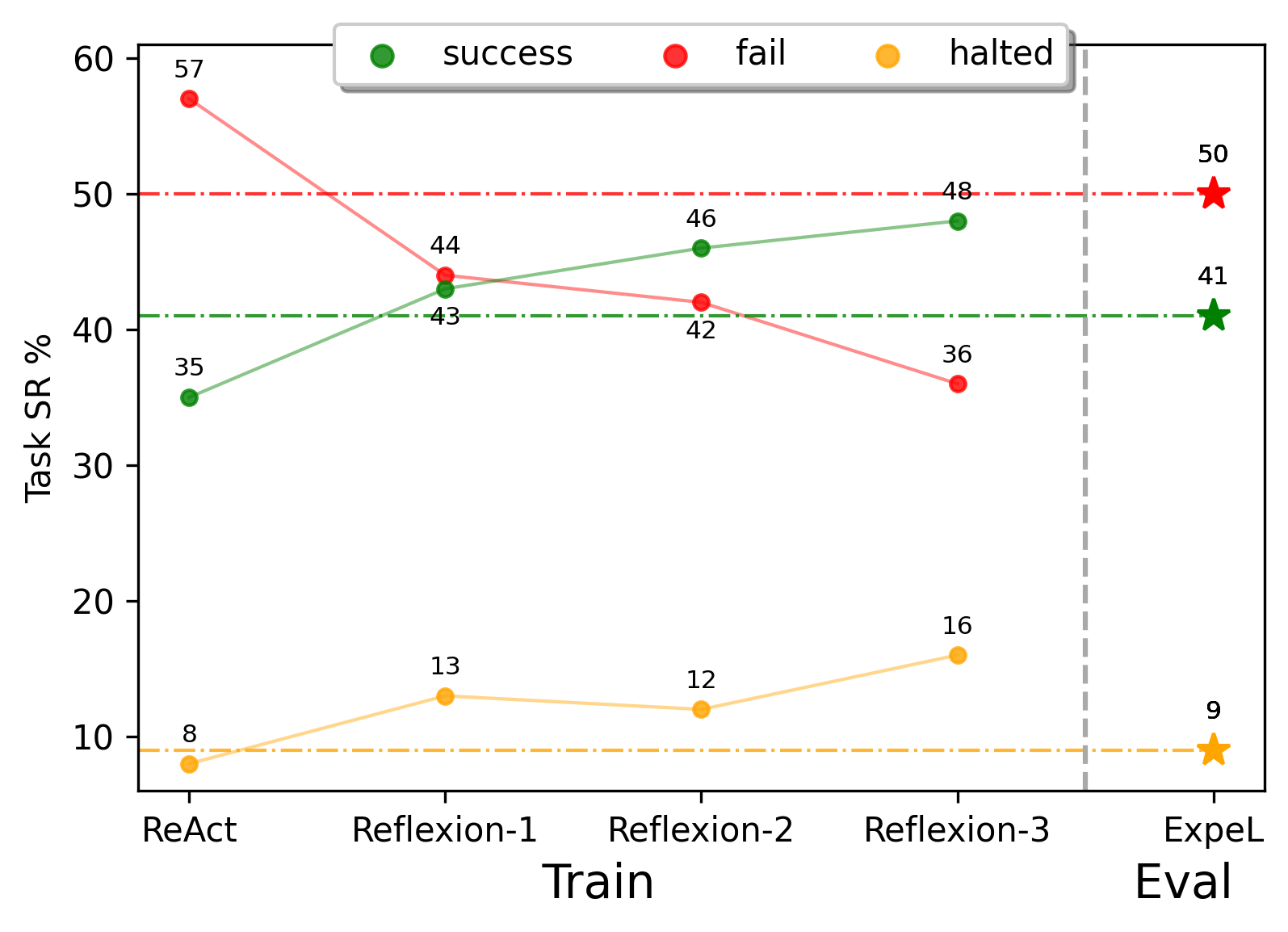}
    \caption{\textbf{Training \& Evaluation outcomes breakdown, WebShop.}}
    \label{fig:webshop_train}
\end{figure*}

We share additional quantitative results in this section. A breakdown of average success rate per environment for ALFWorld and environment-specific average reward for WebShop is presented in Tab. \ref{tab:main-result}. Breakdown of task outcomes (success, failed, halt) is illustrated in Fig. \ref{fig:hotpot_train}, \ref{fig:alf_train}, and \ref{fig:webshop_train}, for HotpotQA, ALFWorld, and WebShop, respectively. Finally, we share some additional metrics regarding step statistics and used tokens in Tab. \ref{tab:add-results}.
\begin{table*}[ht]
\centering
\resizebox{\linewidth}{!}{
    \begin{tabular}{@{}cccccccccc@{}}
    \toprule
    \multicolumn{2}{c}{} & \quad & \multicolumn{1}{c}{\textbf{Gradient-based}} & \quad & \multicolumn{5}{c}{\textbf{Prompt-based}} \\
    \cmidrule{4-4} \cmidrule{6-10}
    \multirow{-2}{*}{\textbf{Benchmark}} & \multirow{-2}{*}{\textbf{Env. Name}} & \quad & Imitation Learning & \quad & Act & ReAct & ExpeL (insights) & ExpeL (retrieve) & ExpeL (ours) \\
    \midrule \rowcolor{MK_Three_One!10}
    \textbf{ALFWorld (SR \%)} & put && 46 && 46 & 50 & 61 & 73 & \textbf{83}\\
                        & clean && 39 && 39 & 61 & \textbf{87} & 74 & 74 \\ \rowcolor{MK_Three_One!10}
                        & heat && \textbf{74} && 4 & 13 & 12 & 43 & 43\\
                        & cool && \textbf{100} && 48 & 71 & 76 & 71 & 67 \\ \rowcolor{MK_Three_One!10}
                        & look && 22 && 11 & 0 & 0 & 17 & \textbf{39} \\
                        & puttwo && 24 && 6 & 0 & \textbf{29} & \textbf{29} & \textbf{29} \\ \rowcolor{MK_Three_One!10}
    \midrule
    \textbf{WebShop ($r$ score)}& shop && 0.599 && 0.666 & 0.665 & 0.675 & 0.67 & \textbf{0.701} \\
    \midrule
    \end{tabular}}
    \caption{\textbf{Environment-Specific Scores.} We present the decomposed ALFWorld success rate per environment name and the WebShop mean environment average reward (See Appendix \ref{sec:webshop_reward_function} for the reward function).}
    \label{tab:main-result}
\end{table*} 

 \FloatBarrier

 \FloatBarrier

\begin{table*}[!ht]
\centering
    \begin{tabular}{@{}llccc@{}}
        \toprule
        \textbf{Additional Metrics (avg. per traj.)} &  & \textbf{HotpotQA} & \textbf{ALFWorld} & \textbf{WebShop} \\
        \midrule
        \multirow{5}{*}{Number of thoughts}
        & Act & 0.0& 0.0& 0.0\\
		& ReAct & 5.19& 8.96& 3.08\\
		& Insights-only & 5.28& 7.57& 3.26\\
		& Retrieve-only & 4.65& 7.9& 2.91\\
		& ExpeL & 5.02& 8.16& 3.2\\
        \midrule
        \multirow{5}{*}{Number of actions} 
		& Act & 5.08& 11.13& 4.32\\
		& ReAct & 5.18& 14.82& 4.47\\
		& Insights-only & 5.04& 14.0& 4.72\\
		& Retrieve-only & 4.63& 13.08& 4.24\\
		& ExpeL & 4.8& 14.3& 4.33\\
        \midrule
        \multirow{5}{*}{Number of observations} 
		& Act & 5.08& 23.37& 4.37\\
		& ReAct & 5.19& 20.01& 7.68\\
		& Insights-only & 5.12& 18.1& 8.05\\
		& Retrieve-only & 4.63& 17.22& 7.55\\
		& ExpeL & 4.87& 18.32& 7.56\\
        \midrule
        \multirow{5}{*}{Number of invalid actions} 
		& Act & 0.0& 6.25& 0.16\\
		& ReAct & 0.0& 2.84& 0.42\\
		& Insights-only & 0.01& 2.34& 0.26\\
		& Retrieve-only & 0.01& 1.95& 0.61\\
		& ExpeL & 0.03& 2.32& 0.35\\
        \midrule
        \multirow{5}{*}{Tokens} 
        & Act & 1920.48& 1498.63& 2191.57\\
		& ReAct & 1319.75& 2051.49& 2575.41\\
		& Insights-only & 3525.7& 2790.05& 3224.95\\
		& Retrieve-only & 3609.43& 2190.35& 2889.57\\
		& ExpeL & 4310.06& 2856.7& 3291.31\\
        \midrule
        \multirow{5}{*}{Thought tokens} 
		& Act & 0.0& 0.0& 0.0\\
		& ReAct & 192.51& 282.28& 116.41\\
		& Insights-only & 231.48& 241.62& 118.8\\
		& Retrieve-only & 176.71& 260.27& 103.52\\
		& ExpeL & 212.13& 262.66& 111.51\\
        \midrule
        \multirow{5}{*}{Action tokens} 
		& Act & 58.79& 81.19& 43.8\\
		& ReAct & 68.07& 104.14& 45.33\\
		& Insights-only & 71.4& 98.98& 50.39\\
		& Retrieve-only & 60.34& 93.75& 44.35\\
		& ExpeL & 66.41& 100.78& 44.99\\
        \midrule
        \multirow{5}{*}{Observation tokens} 
		& Act & 445.72& 416.46& 41.52\\
		& ReAct & 625.46& 393.16& 58.27\\
		& Insights-only & 560.42& 384.54& 58.97\\
		& Retrieve-only & 496.69& 376.1& 56.66\\
		& ExpeL & 547.23& 393.19& 57.23\\
        \midrule
    \end{tabular}
    \bigskip
    \caption{\textbf{Additional Statistical Metrics.} Average counts per trajectory for each benchmark. All strings were tokenized using \texttt{tiktoken} (\url{https://github.com/openai/tiktoken}).}
    \label{tab:add-results}
\end{table*}

\end{document}